\documentclass[]{interact}
\usepackage{amssymb}
\usepackage{amsmath}
\usepackage{hyperref}
\hypersetup{
    colorlinks=true,
    linkcolor=cyan,
    filecolor=cyan,
    urlcolor=cyan,
    citecolor=cyan,
}
\usepackage{natbib}% Citation support using natbib.sty
\usepackage{url}
\usepackage{epstopdf}% To incorporate .eps illustrations using PDFLaTeX, etc.
\usepackage{subfigure}% Support for small, `sub' figures and tables
\usepackage{multirow}
\usepackage{natbib}% Citation support using natbib.sty
\bibpunct[, ]{(}{)}{,}{a}{}{,}% Citation support using natbib.sty
\renewcommand\bibfont{\fontsize{10}{12}\selectfont}% Bibliography support using natbib.sty

\theoremstyle{plain}% Theorem-like structures

\usepackage{CJKutf8}
\theoremstyle{definition}

\usepackage{algorithm}
\usepackage{algorithmic}
\usepackage{amsfonts}
\usepackage{amsmath}
\usepackage{multirow}
\usepackage{tabularx}   % For flexible widths
\usepackage{xcolor}
\usepackage{pifont}
\theoremstyle{remark}

\usepackage{array}
\usepackage{multirow}
\usepackage{adjustbox}
\usepackage{amssymb}
\usepackage{lmodern}
\usepackage{comment}

\begin{document}

 \articletype{ARTICLE}

\title{A Decade of PhotoMappers: A Longitudinal Study of Crowdsourced Disaster Photos Geolocalization with Vision-Language Models and Geospatial Reasoning}

\author{
\name{
Hao Li\textsuperscript{a*},
Wenping Yin\textsuperscript{b},
Fabian Deuser\textsuperscript{c},
Jia Jia\textsuperscript{a},
Ziqi Liu\textsuperscript{d},
Levente Juhász\textsuperscript{e,f},
Fan Zhang\textsuperscript{g},
and Filip Biljecki\textsuperscript{h,i}
\thanks{CONTACT Hao Li. Email: hao.li@nus.edu.sg}}
\affil{
\textsuperscript{a} Department of Geography, National University of Singapore, Singapore; 
\textsuperscript{b} College of Geodesy and Geomatics, Shandong University of Science and Technology, China;
\textsuperscript{c} University of the Bundeswehr Munich, Germany;
\textsuperscript{d} State Key Laboratory of Information Engineering in Surveying, Mapping and Remote Sensing, Wuhan University, China;
\textsuperscript{e} School of Forest, Fisheries and Geomatics Sciences, University of Florida, Gainesville, FL, USA;
\textsuperscript{f} Fort Lauderdale Research and Education Center, University of Florida, Davie, FL, USA;
\textsuperscript{g} Institute of Remote Sensing and GIS, Peking University, China;
\textsuperscript{h} Department of Architecture, National University of Singapore, Singapore; 
\textsuperscript{i} Department of Real Estate, National University of Singapore, Singapore}
}

\maketitle

\begin{abstract}

Crowdsourced disaster imagery provides timely, ground-level observations when conventional disaster mapping is delayed or incomplete, yet its geographic value depends on where observations are produced, how they are interpreted, and whether they can be reliably located as actionable Geospatial Intelligence (GEOINT). Using more than a decade of PhotoMappers, a crowdsourced disaster-mapping initiative, this study examines the changing geography of disaster information and asks how emerging GeoAI models compare with human geographic interpretation. We combine longitudinal spatial analysis with cross-view geolocalization using street view imagery (SVI) and remote sensing imagery (RSI), and evaluate vision-language models (VLMs) reasoning through 30 indicators spanning geographic cues and reasoning evidence. The results reveal a geographically and temporally uneven archive shaped by changing disaster types, Community Lifelines, and operational priorities. Integrating VGI, SVI, and RSI with Sinkhorn re-ranking achieves 40.90\% R@1, 61.10\% R@5, and 69.45\% R@10, while yearly R@1 varies markedly from 25.00\% to 65.07\%. Geographic reasoning is similarly uneven: Qwen3.6-27B achieves an overall agreement of 67\% compared with 51\% for InternVL3-8B, with both models showing greater difficulty with localized and place-specific evidence than broad spatial configurations. These findings show that geolocalization and geographic reasoning are related but distinct capabilities, shifting attention from whether AI can autonomously geolocate disaster imagery to when, where, and through what geographic evidence it can do so reliably. We argue that the value of GeoAI lies less in replacing volunteered geographic interpretation than in complementing it, scaling candidate retrieval while retaining human judgment for ambiguous and consequential cases. This provides a pathway toward more human-centered and responsible GeoAI for disaster mapping. All data and code are openly available at \url{https://github.com/bobleegogogo/PhotoMappers-VLM}.

\end{abstract}


\section{Introduction}

Climate change, rapid urbanization, and growing exposure are increasing the risks and consequences of weather- and climate-related disasters. Recent hurricanes, wildfires, and urban floods illustrate a persistent challenge for disaster response, as conditions on the ground can change rapidly \citep{noaa2024helene}. Decision-makers require timely and geographically precise information about impacts on the ground to protect lives, direct emergency services, and minimize damage. The World Meteorological Organization reported that annual global mean near-surface temperatures during 2026–2030 will range between 1.3°C and 1.9°C above the 1850-1900 average, with an 86\% probability that at least one year during this period will surpass 2024 as the warmest year on record \citep{wmo2026}. Recent studies have further linked the severity of some recent events to climate change, for example, NOAA reported that climate change increased rainfall associated with Hurricane Helene \citep{noaa2024helene}. In this context, there is a pressing need for rapid, localized, and reliable geographic information during disasters.

Crowdsourced imagery and Volunteered Geographic Information (VGI) have been important complements to conventional disaster observations. Citizens, first responders, journalists, and other observers generate large volumes of photographs documenting flooded roads, damaged buildings, wildfires, debris, and disrupted infrastructure. Such VGI disaster photos provide timely street-level evidence that can complement satellite and authoritative data, particularly where rapid change, cloud cover, or limited observational capacity constrains conventional data collection. VGI and crowdsourcing have consequently become increasingly integrated into disaster mapping and disaster response workflows \citep{goodchild2007,goodchild2010,haworth2015,haworth2018}. Yet a photograph becomes actionable geographic information only when its content can be interpreted, located, and validated with sufficient confidence. Beyond traditional VGI, these crowdsourced disaster observations are also increasingly recognized by US federal and emergency management agencies, including the Federal Emergency Management Agency (FEMA) and the National Geospatial-Intelligence Agency (NGA), as critical form of Open-Source Intelligence (OSINT). When unstructured social media and ground-level photographs are verified and spatially located, they transform into actionable Geospatial Intelligence (GEOINT) \citep{fleming_ethics_2018}.  However, utilizing unstructured open-source GEOINT requires rapid, reliable geolocalization. The volume and velocity of crowdsourced data, including imagery, exceed manual processing capabilities, which creates an operational bottleneck that human analysts alone struggle to scale during disaster response \citep{imran_processing_2015, haworth2018, huck_centaur_2021}.

The PhotoMappers project provides a unique operational example of utilizing VGI in disaster response. Since 2017, PhotoMappers has mobilized volunteers to identify and geolocate disaster-related photographs from social media and other public sources for emergency-management applications \citep{giscorps2026}. Volunteers are asked to infer locations from geographic cues such as roads, signs, buildings, landmarks, infrastructure, and landscape characteristics, after which submissions are vetted and enriched with disaster-relevant attributes. During Hurricane Ian in 2022, for example, volunteers geolocated 457 images within three days \citep{giscorps2022}. PhotoMappers therefore represents more than a crowdsourced image archive, serving as a sustained mapping activity through which human visual interpretation transforms unstructured observations into validated geographic information.

Recent advances in large-language models (LLMs), especially vision-language models (VLMs), provide the possibility of automating or augmenting this crowdsourced mapping process. By jointly reasoning over visual and textual information, VLMs can recognize scenes and objects, interpret text, integrate contextual cues, and infer geographic locations. Recent studies have demonstrated their emerging capabilities in visual geolocalization and geospatial reasoning \citep{li2024georeasoner,li2025cross, danish2025geobenchvlm, ji2025geospatial}. These advanced capabilities of VLMs are particularly relevant to crowdsourced disaster imagery geolocalization, where location is often inferred from combinations of weak geographic cues, including road networks, business names, architecture, vegetation, infrastructure, and landscape morphology, rather than from a single definitive landmark.

However, general-purpose multimodal capability of VLMs does not necessarily translate into reliable geographic reasoning. Disaster imagery is particularly challenging because scenes may be damaged, visually ambiguous, temporally dynamic, or obscured by floodwater, smoke, and debris. VLMs may correctly recognize scene elements while drawing incorrect geographic inferences, relying on spurious contextual associations, or producing over-confident but unsupported interpretations. Recent benchmarks similarly show that strong general-purpose multimodal performance does not guarantee robust geospatial reasoning \citep{danish2025geobenchvlm,ji2025geospatial,hochmair2025advancing}. Such errors are crucial in disaster response, where incorrect locations can propagate into downstream mapping and decision-making. Therefore, it becomes a critical task for responsible Geospatial Artificial Intelligence (GeoAI) models to optimize not only for geolocalization accuracy but also for the geographic reasoning behind model predictions.

\begin{figure*}[!ht]
    \centering
    \includegraphics[width=\linewidth]{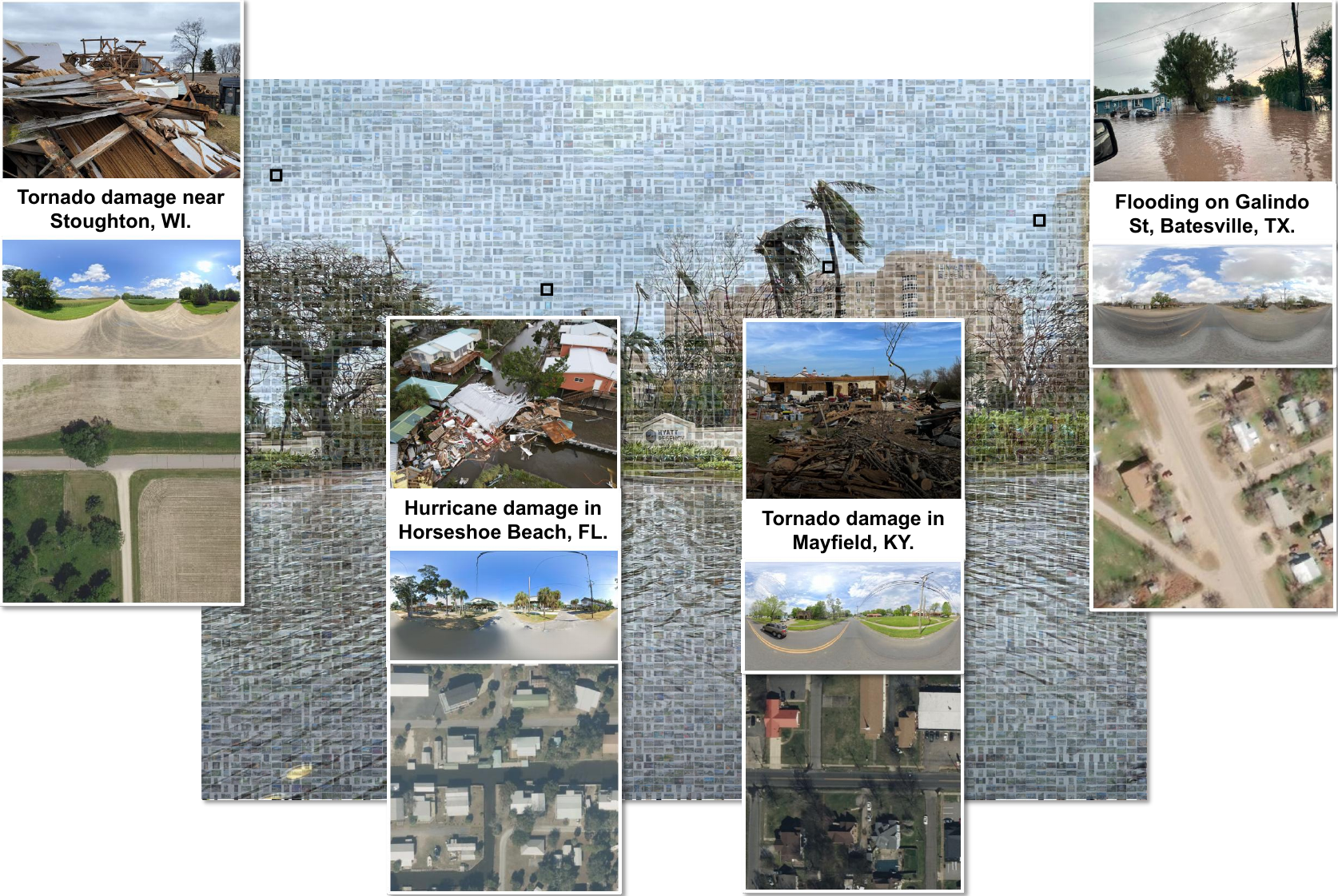}
    \caption{Examples of crowdsourced VGI from the PhotoMappers dataset across different disaster types. The background is a mosaic of over 8,000 VGI images; four representative examples show VGI, SVI, and RSI.}
    \label{fig:dataset}
\end{figure*}

In this paper, we curated and extended the decade-long PhotoMappers archive (i.e., from 2017 to 2026) into a cross-view geolocalization dataset with text-based description, in order to examine how crowdsourced disaster mapping has evolved and how contemporary VLMs perform on the same class of real-world geographic observations. 
Figure~\ref{fig:dataset} illustrates representative examples from the resulting PhotoMappers archive, where examples demonstrate the substantial visual heterogeneity of the dataset, including differences in disaster phenomena, built environments, image quality, viewing angle, and availability of explicit geographic clues. Such heterogeneity is central to our objective of evaluating VLMs under real-world disaster scenarios. Rather than treating the curated PhotoMappers dataset solely as a benchmark for cross-view geolocalization task, we approach it as a longitudinal record of geographic information production across disasters, places, and years. Specifically, we first examine yearly changes in the geographic distribution and spatial concentration of PhotoMappers observations. Then, we further evaluate geolocalization using complementary street-view imagery (SVI) and remote-sensing imagery (RSI), followed by an indicator-based assessment of the geographic evidence used by VLMs to explain cross-view correspondence. In a nutshell, this paper addresses three research questions:

\begin{itemize}

    \item \textbf{RQ1:} How has the geography of crowdsourced PhotoMappers archive evolved over time?
    \item \textbf{RQ2:} To which extent can GeoAI achieve human-level geolocalization of crowdsourced disaster imagery?
    \item \textbf{RQ3:} How do VLMs and humans differ in geospatial reasoning with disaster photos?

\end{itemize}

Furthermore, the longitudinal analysis in this paper moves beyond geolocalization accuracy to examine how geographic evidence is interpreted and validated by humans and AI. In doing so, the study identifies where VLMs can provide accurate interpretation, where human geographic judgment remains essential, and how Human-AI interaction may support more reliable and responsible crowdsourced disaster mapping in the future.

%----------------------------------------------
%----------------------------------------------

\section{Related Work}

\subsection{VGI and Crowdsourcing in Disaster Response}

VGI and crowdsourcing have transformed how geographic information is produced and mobilized during disasters. \citet{goodchild2007} conceptualized VGI as distributed geographic sensing through which citizens contribute observations of places and events. Its capacity for rapid and localized information production has made VGI particularly valuable for disaster mapping, situational awareness, and emergency response when authoritative information is incomplete or delayed~\citep{goodchild2010,haworth2015}. At the same time, VGI raises persistent questions of quality, credibility, representativeness, and validation because participation and information availability vary across contributors, places, and events~\citep{flanagin2008,haklay2010,fan2014quality}.

Moreover, it is often emphasized that VGI should be understood as a social and institutional process of knowledge production rather than simply a large collection of observations. \citet{haworth2018}, for example, highlighted the roles of scale, trust, uneven participation, and the relationship between citizen-generated and authoritative geographic information in disaster management. \citet{burns2018} similarly showed that digital humanitarian mapping can fill gaps in formal disaster response while simultaneously shaping which needs, places, and communities become represented as data. \citet{Herfort2019} demonstrated that combining crowdsourcing with deep learning can improve human-settlement mapping while reducing volunteer effort, illustrating how machine learning can augment rather than simply replace human geographic contributions. \citet{huck_centaur_2021} extended this idea through Centaur VGI, explicitly combining human spatial cognition with machine efficiency to address persistent inequalities in map coverage. These studies position VGI as a unique and dynamic geographic process in which participation, representation, validation, and increasingly human--machine collaboration shape what becomes mapped and operationally visible.

Crowdsourced photographs extend VGI by providing visual evidence of roads, buildings, infrastructure, signs, landscapes, and disaster impacts \citep{juhasz_user_2016}, which then can subsequently be used for mapping \citep{juhasz2016cross}. Existing disaster-image datasets have primarily supported event classification, damage assessment, and situational awareness~\citep{alam2018crisismmd}, with less attention to how geographic location itself is inferred from visual evidence. PhotoMappers is distinctive in this respect, in which volunteers interpret and geolocate disaster photographs before their submissions are vetted and incorporated into operational PhotoMappers products~\citep{giscorps2026}. Moreover, its multi-year archive provides further an opportunity to examine not only individual observations, but also how the longitudinal changes of crowdsourced disaster mapping has evolved across places, events, and time.

\subsection{Crowdsourced Imagery Geolocalization with GeoAI}

Image geolocalization has conventionally been approached as a visual recognition or retrieval task, linking query images to geographically referenced imagery (e.g., RSI) through jointly learned visual representations. Such approaches perform particularly well where stable and distinctive streetscapes or landscape features are available. Crowdsourced disaster imagery presents a more difficult geographic setting because scenes may be altered or obscured by flooding, debris, smoke, and structural damage, while useful visual cues can be limited.

Recent GeoAI research has increasingly addressed this challenge through cross-view and multimodal geographic information fusion. \citet{li2025cross} introduced CVDisaster, integrating street-view and very-high-resolution satellite imagery to jointly support cross-view geolocalization and damage perception for Hurricane Ian. This work demonstrated the value of connecting ground-level and overhead geographic contexts, but remained centered on structured cross-view imagery such as SVI and RSI. \citet{yin2026triple} extended this direction to disaster-related VGI by integrating crowdsourced images and text with SVI and RSI. Their MultiIan framework employed LLMs to extract implicit geographic information from VGI text and triple-objective cross-view learning to connect VGI, SVI, and RSI, demonstrating the potential of multimodal geographic context for localizing heterogeneous disaster observations.

More recent studies have begun to move beyond deterministic cross-view retrieval. \citet{li2026generative} proposed ProbGLC, which combines deterministic and probabilistic cross-view geolocalization to represent candidate locations as geographic probability distributions and quantify their localizability. This shifts disaster geolocalization from predicting a single location toward explicitly representing spatial uncertainty, which is particularly relevant when disaster imagery contains weak or ambiguous geographic evidence. Recently, \citet{yin2026disastertd} introduced DisasterTD to combine multimodal LLMs with cross-view geolocalization for disaster toponym disambiguation, extending location inference from visual matching toward the joint interpretation of textual place references and multimodal geographic context. Related work has also explored language-model-guided attention for identifying geographically informative regions within dynamic crowdsourced flood imagery~\citep{xu2026flood}, alongside broader developments in multimodal disaster GeoAI~\citep{li2026synergizing}.

These technical advances intersect with a broader geographic concern: disaster-related VGI is produced unevenly across places, events, and communities. Research in \textit{Annals} has shown that social-media observations can capture spatial and temporal patterns of disaster activity and contribute to post-disaster damage estimation, while simultaneously revealing substantial geographic disparities in participation and information availability~\citep{zou2018,zhou2026rapid}. Cross-view, probabilistic, and multimodal GeoAI can expand the geographic information extracted from crowdsourced observations, but their performance remains embedded within the uneven spatial and contextual conditions under which disaster VGI is produced.

%Furthermore, existing geolocalization studies remain predominantly oriented toward localization accuracy, commonly measured through retrieval metrics or geographic distance. Recent probabilistic approaches have begun to expose uncertainty in where an image may have been captured, but these metrics and distributions still reveal less about the geographic evidence underlying a prediction. A correct location may be inferred from a landmark, spatial configuration, textual clue, contextual association, or several weak geographic cues; conversely, an incorrect prediction may reflect ambiguous or conflicting evidence rather than an absence of meaningful geographic reasoning.

For disaster response, the key question is therefore not only how accurately GeoAI can locate an image, but also how geographic evidence is interpreted across views and whether that interpretation corresponds to the reasoning through which human mappers establish location. This concern motivates the critical shift in this study from cross-view geolocalization alone toward the explicit evaluation of geospatial reasoning.

\subsection{Vision-Language Models and Geospatial Reasoning}

VLMs extend image analysis from detection and recognition toward multimodal interpretation and reasoning. By integrating visual information, language, and contextual knowledge, VLMs show a great potential to recognize geographic landmarks, interpret embedded text, describe spatial relationships, and generate explanations for their predictions. These capabilities have stimulated growing interest in visual geolocalization and broader geospatial reasoning tasks in the GIScience community \citep{li2024georeasoner,danish2025geobenchvlm,ji2025geospatial}.

Recent studies, however, suggest that geographic reasoning remains a challenging task for general-purpose VLMs. GeoReasoner~\citep{li2024georeasoner} demonstrates that incorporating explicit human geographic reasoning can improve visual geolocalization beyond sole image-location matching, while recent benchmarks and geospatial foundation models (GeoFMs) reveal persistent limitations in understanding e.g., geographic entities ~\citep{danish2025geobenchvlm}, spatial relations ~\citep{ji2025geospatial}, and geographical context ~\citep{dorobantu2026geospatial}. For disaster imagery, this reasoning process may involve integrating street signs, built-environment characteristics, road configurations, vegetation, infrastructure, and landscape features. VLMs offer a mechanism for such multimodal reasoning, but their inferences may also reflect biased or geographically uneven learned associations and remain vulnerable to unsupported reasoning or hallucination, especially in the time-critical disaster response application~\citep{2026_npj_bias}.

This recent trend is particularly important from a geographic perspective because VLM's reasoning strategy is not necessarily spatially neutral. Spatial biases and heterogeneity in training data, visual representation, and geographic context may shape what places and features VLMs recognize and how confidently they reason about them~\citep{2026_npj_bias}. At the same time, hybrid human-machine approaches suggest that AI automation can not fully replace volunteer mapping. For example, Centaur VGI \citep{watkinson2022centaur} demonstrates how machine learning can improve mapping speed and data quality while retaining purposeful human participation. As a consequence, these consideration shifts attention from whether AI (e.g., LLMs or VLMs) can replace human geographic interpretation toward how human and machine reasoning may complement, reinforce, or diverge from one another.

\section{A Decade of PhotoMappers}

\subsection{The PhotoMappers Project and Operational workflow}

PhotoMappers is a decade-long crowdsourcing project and disaster-mapping initiative, which was developed through the partnership of the National Alliance for Public Safety GIS (NAPSG) Foundation and GISCorps, with CEDR Digital Corps \footnote{https://cedrdigitalcorps.org/} subsequently contributing social-media search capabilities and technical support. PhotoMappers traces its origins to NAPSG-supported crowdsourcing missions conducted during the 2017 hurricane season, when GISCorps volunteers were mobilized to locate and geolocate crowdsourced photographs associated with Hurricanes Harvey, Irma, and Maria~\citep{giscorps2017}. Since then, the initiative has evolved from individual disaster-response missions into an operational workflow covering hurricanes, floods, tornadoes, earthquakes, wildfires, and other climate extreme events across the United States and its territories. The official project archive identifies missions across every year from 2017 onward, and as of 2026, the PhotoMappers project is in its tenth year of providing situational awareness to federal, state, and local emergency managers~\citep{giscorps2026}. 

\begin{figure*}[!ht]
    \centering
    \includegraphics[width=\linewidth]{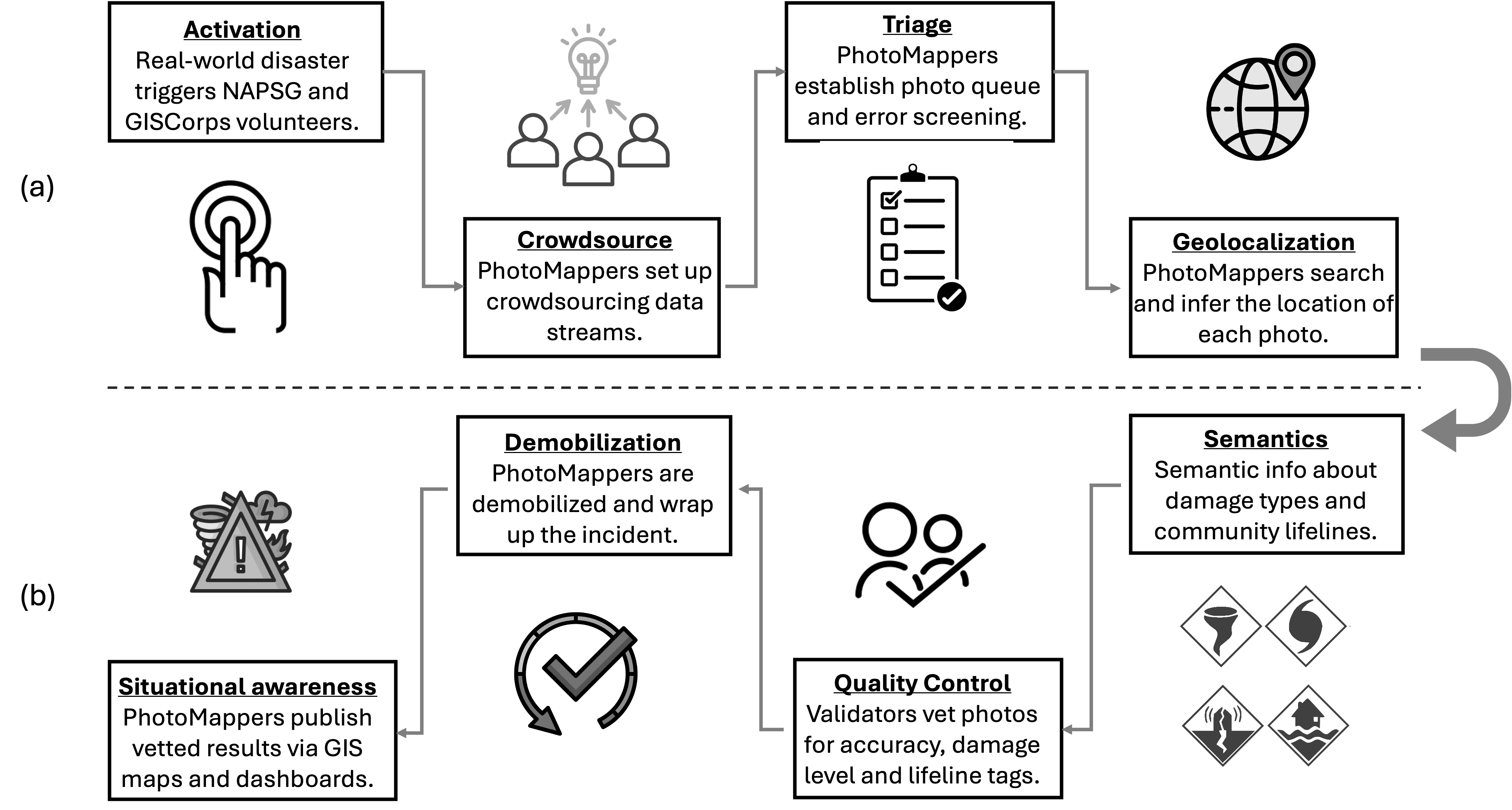}
    \caption{The practical workflow of PhotoMappers, comprising a) activation, crowdsourcing, triage, geolocalization and b) semantic enrichment, quality control, demobilization, and situational awareness. The workflow illustrates how crowdsourced disaster photographs are transformed from distributed observations into validated geographic information for disaster response.}
    \label{fig:workflow}
\end{figure*}

The core task of PhotoMappers is to convert publicly available photographs into timely, geographically explicit information that can complement conventional disaster observations, as shown in Figure~\ref{fig:workflow}. During a PhotoMappers campaign, NAPSG first coordinates with emergency-management partners and identifies areas or issues requiring attention. GISCorps then mobilizes volunteers who search social-media platforms and news sources for relevant photographs of on-the-ground conditions. Next, volunteers come in and interpret visual and contextual clues, including road infrastructure, signage, buildings, businesses, landmarks, vegetation, terrain, and accompanying text, to determine the most plausible location of each photograph. As a result, the inferred geolocation is submitted through a structured web-based workflow, where the photograph and associated attributes become part of the PhotoMappers dataset ~\citep{giscorps2022,giscorps2023}.

%----------------------------------------------
%----------------------------------------------

As illustrated in Figure~\ref{fig:workflow}, the operational workflow of PhotoMappers comprises roughly eight stages: activation, crowdsourcing, triage, geolocalization, semantic enrichment, quality control, demobilization, and situational awareness. An \textbf{activation} begins when NAPSG and emergency-management partners identify an event and associated information needs. During \textbf{crowdsourcing}, GISCorps volunteers and supporting tools search social media, news, and other public sources for relevant disaster photographs, which are subsequently \textbf{triaged} for relevance and usability. During \textbf{geolocalization}, volunteers infer photograph locations from visual and contextual evidence, including roads, signs, buildings, businesses, landmarks, vegetation, terrain, and accompanying text. The records are then \textbf{semantically enriched} with disaster-relevant information, including observed impacts and, where applicable, FEMA Community Lifeline categories and damage information. \textbf{Quality control} by the PhotoMappers Admin Team reviews geographic accuracy, relevance, and associated attributes before validated observations are incorporated into operational products. As information needs decline, \textbf{demobilization} concludes the activation and preserves the resulting records within the project archive. Finally, validated observations support \textbf{situational awareness} through maps, dashboards, and other information products used by emergency managers and first responders~\citep{giscorps2021,giscorps2022,giscorps2023,giscorps2026}. Importantly, these stages already constitute a human-in-the-loop process through which distributed visual observations are transformed into validated and actionable geographic information.

%The workflow has progressively combined volunteer geographic interpretation with geospatial technologies and automation. Early activations relied largely on manual social-media searches, whereas automated image discovery through CEDR Digital Corps was introduced by 2021~\citep{giscorps2021}. By 2022--2024, PhotoMappers had developed a more integrated workflow combining automated social-media discovery, volunteer geolocalization through Survey123 and Esri Experience Builder, and systematic administrative validation~\citep{giscorps2022,giscorps2023,giscorps2024}. PhotoMappers thus provides a longitudinal record of the evolving relationship among crowdsourcing, automation, and human geographic judgment in disaster information production.

The operational significance of this workflow is evident across major activations. PhotoMappers mapped and vetted more than 170 photographs within 36 hours during the 2021 Quad State Tornadoes ~\citep{giscorps2021}, 457 photographs over three days during Hurricane Ian in 2022 ~\citep{giscorps2022}, and 677 photographs during Hurricane Helene in 2024 ~\citep{giscorps2024}. During Helene, persistent cloud cover further constrained aerial observations, increasing the value of crowdsourced ground-level imagery. These activations demonstrate both the capacity of crowdsourcing to rapidly produce actionable geographic information and its continued dependence on labor-intensive human interpretation and validation, which largely motivates our examination of GeoAI and VLMs.

\subsection{Defining and Integrating the Community Lifelines}

Defined by Federal Emergency Management Agency (FEMA), Community Lifelines, as listed in Table \ref{tab:community_lifelines}, provide an operational framework for translating disaster observations into information relevant to emergency response. As introduced in the 4th Edition National Response Framework (NRF), in total eight Community Lifelines, including Safety and Security; Food, Water, Shelter; Health and Medical; Energy; Communications; Transportation; and Hazardous Material, represent critical services essential to human health, safety, and economic security. By organizing observations around disrupted community functions and their inter-dependencies, the Community Lifeline framework provides a common language for assessing impacts, prioritizing response, and monitoring changing conditions~\citep{fema2019nrf}. For PhotoMappers, it provides a semantic bridge between what is visible in a photograph and why that observation matters for disaster response.

%----------------------------------------------
\begin{table*}[!ht]
\centering
\caption{FEMA Community Lifelines definitions and their primary components.}
\label{tab:community_lifelines}

\small
\renewcommand{\arraystretch}{1.15}
\setlength{\tabcolsep}{4pt}

\begin{adjustbox}{max width=\textwidth,center}
\begin{tabular}{
>{\raggedright\arraybackslash}m{0.22\textwidth}
>{\centering\arraybackslash}m{0.10\textwidth}
>{\raggedright\arraybackslash}m{0.62\textwidth}
}
\hline
\textbf{Community Lifeline} &
\textbf{} &
\textbf{Definition and Primary Components} \\
\hline

\textbf{Safety and Security} &
\includegraphics[height=1.05cm]{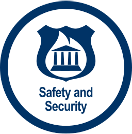} &
Law enforcement and government services supporting communal security,
search and rescue, evacuation, firefighting, and responder safety. \\

\textbf{Food, Hydration, Shelter} &
\includegraphics[height=1.05cm]{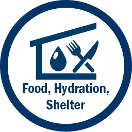} &
Services sustaining basic human needs, including food, hydration, shelter,
and agriculture. \\

\textbf{Health and Medical} &
\includegraphics[height=1.05cm]{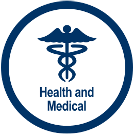} &
Infrastructure and services supporting medical care, public health, patient
movement, medical supply chains, and fatality management. \\

\textbf{Energy (Power \& Fuel)} &
\includegraphics[height=1.05cm]{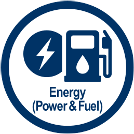} &
Electric power generation, transmission, and distribution, together with
gas and liquid-fuel processing, transportation, and delivery. \\

\textbf{Communications} &
\includegraphics[height=1.05cm]{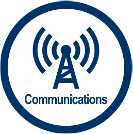} &
Communications infrastructure and services, including broadband, cellular,
broadcast, emergency communications, alerts and warnings, 911, and dispatch. \\

\textbf{Transportation} &
\includegraphics[height=1.05cm]{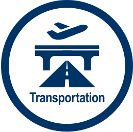} &
Transportation systems and infrastructure, including roadways, mass transit,
railways, aviation, maritime, pipeline, and intermodal networks. \\

\textbf{Hazardous Materials} &
\includegraphics[height=1.05cm]{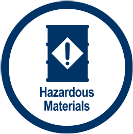} &
Systems addressing hazardous-material facilities, pollutants, contaminants,
incident debris, and other threats to public health and the environment. \\

\textbf{Water Systems} &
\includegraphics[height=1.05cm]{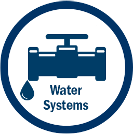} &
Infrastructure and services supporting potable water supply, treatment and
distribution, and wastewater collection and treatment. \\

\hline
\end{tabular}
\end{adjustbox}

\end{table*}
%----------------------------------------------

%----------------------------------------------
\begin{figure*}[!ht]
    \centering
    \includegraphics[width=\linewidth]{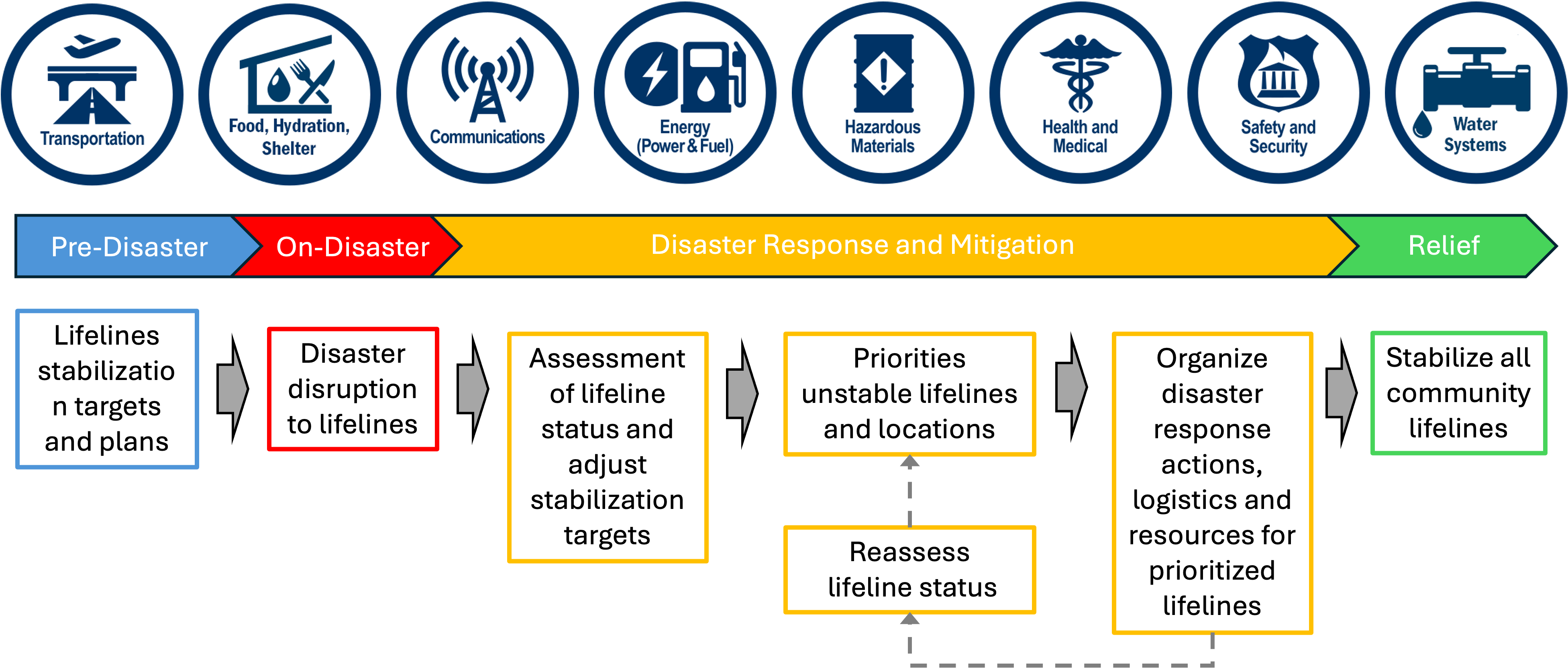}
    \caption{Integration of FEMA Community Lifelines across the PhotoMappers disaster-management workflow. Lifelines connect observed disaster impacts to critical community functions from pre-disaster planning and on-disaster assessment through response, mitigation, and relief.}
    \label{fig:lifelines}
\end{figure*}
%----------------------------------------------

The concept of Community Lifelines is integrated across different stages of the PhotoMappers workflow (as shown in Figure~\ref{fig:lifelines}). For \textbf{pre-disaster}, operational needs help define the geographic and thematic focus of image discovery. For \textbf{on-disaster}, photographs are connected in real-time to affected places and lifeline infrastructures. During \textbf{disaster response and mitigation}, observations are interpreted in relation to disaster impacts and relevant Lifelines. The PhotoMappers Admin Team then performs quality control, reviewing image relevance, geographic accuracy, and associated attributes and, where appropriate, assigning a primary FEMA Community Lifeline and preliminary damage score. As a result, validated observations are subsequently disseminated through maps, dashboards, and the PhotoMappers Hub to support \textbf{disaster relief} via providing situational awareness ~\citep{giscorps2021,giscorps2026}.

\begin{comment}
    
Table~\ref{tab:photomappers_milestones} summarizes the major stages of this development. Rather than treating each year as an isolated dataset release, we view these milestones as an evolution of the project's \textbf{geographic knowledge-production infrastructure}.

\begin{table*}[htbp]
\centering
\caption{Key milestones in the evolution of the PhotoMappers project.}
\label{tab:photomappers_milestones}
\small
\renewcommand{\arraystretch}{1.15}
\begin{tabular}{p{2.5cm} p{3.0cm} p{7.5cm}}
\hline
\textbf{Year} & \textbf{Stage} & \textbf{Key milestone} \\
\hline

2017 &
Initial deployment &
Crowdsourced photo mapping initiated for Hurricanes Harvey, Irma, and Maria. \\

2018--2019 &
Expansion &
Recurring activations established across major hurricane events. \\

2020 &
Multi-hazard mapping &
Coverage expanded to earthquakes, tornadoes, floods, and severe weather. \\

2021 &
Automated discovery &
Automated social-media discovery introduced alongside volunteer mapping and validation. \\

2022 &
Workflow integration &
Photo discovery, volunteer geolocation, validation, and visualization further integrated. \\

2023--2024 &
Operational maturation &
Multi-hazard activations expanded, including wildfires, floods, tornadoes, and hurricanes. \\

2025 &
Decade milestone &
PhotoMappers entered its tenth year of sustained disaster-mapping operations. \\

2026 &
Eleventh year &
Operations continued across increasingly diverse hazards and geographic settings. \\

\hline
\end{tabular}
\end{table*}

\end{comment}

\subsection{Cross-view Dataset Construction}

\begin{figure}[!ht]
    \centering
    \includegraphics[width=\linewidth]{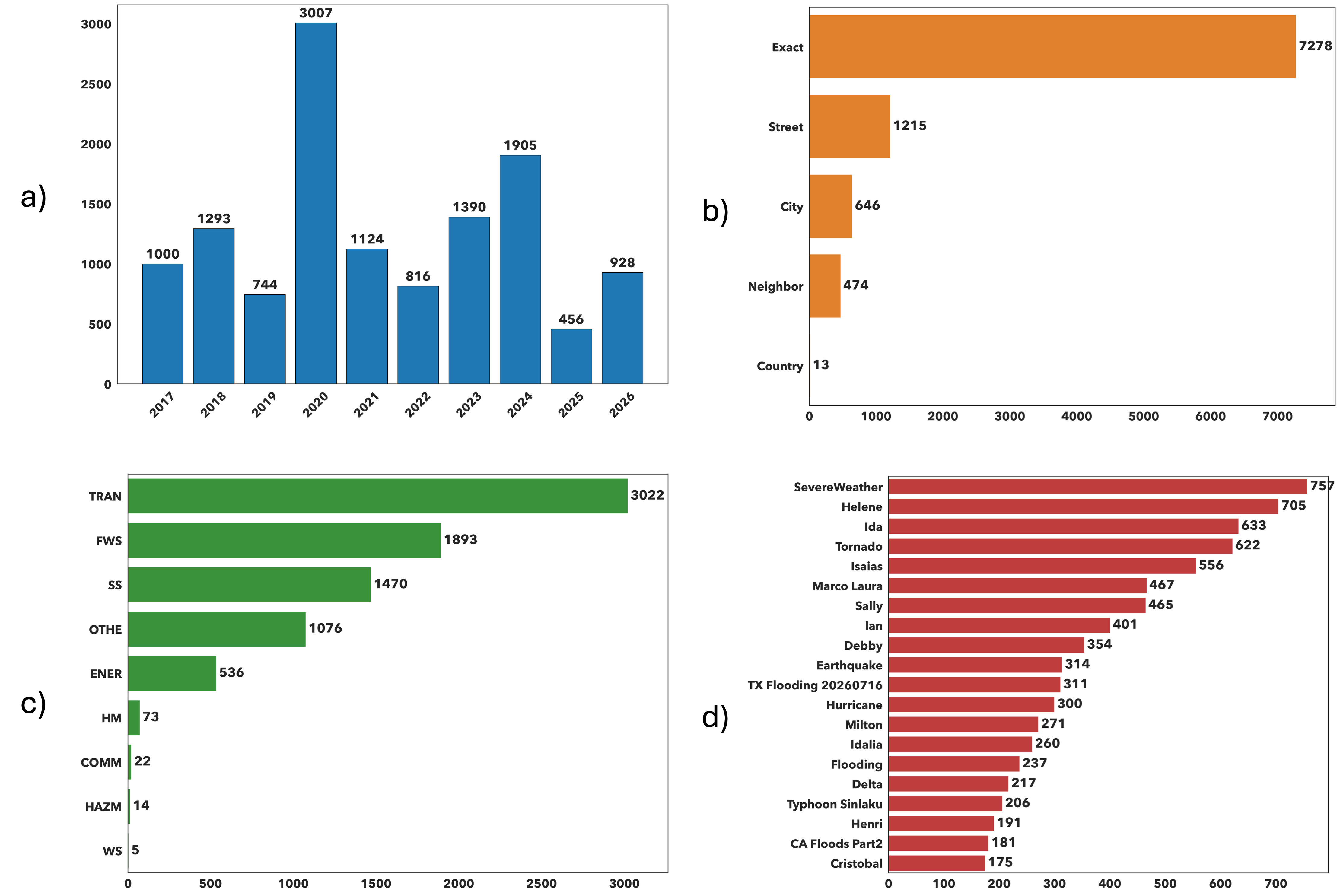}
    \caption{Statistics of selected attributes in the PhotoMappers VGI dataset: (a) annual distribution; (b) geolocation precision; (c) community lifeline categories; and (d) top 20 user-generated disaster tags.}
    \label{fig:data_statistics}
\end{figure}

The original collection comprised 28,502 crowdsourced disaster-image records compiled from multiple PhotoMappers sources and dashboard exports. To reduce spatial redundancy, records with obvious duplicate or highly similar geographic coordinates were removed, yielding 14,440 candidate VGI records. Duplicate image filenames were subsequently removed, resulting in 12,663 unique VGI records as shown in Figure \ref{fig:data_statistics}. Each record was then paired with corresponding SVI and RSI, producing 12,663 cross-view VGI--SVI--RSI image pairs.

Unlike existing cross-view geolocalization benchmarks, these PhotoMappers records originate from a real-world operational disaster-mapping workflow in which photograph locations were inferred by volunteers and subsequently subject to human validation. The resulting coordinates therefore represent human geographic judgments against which GeoAI predictions and reasoning can be evaluated. The paired SVI and RSI further provide complementary ground-level and overhead geographic contexts for assessing cross-view feature alignment.

%----------------------------------------------
%----------------------------------------------

\begin{table}[!ht]
\centering
\caption{Different types of disasters together with representative keywords and record counts.}
\label{tab:disaster_category_annotation}
\small
\renewcommand{\arraystretch}{1.15}
\begin{tabular}{p{3.0cm} p{8.3cm} r}
\hline
\textbf{Disaster Type} &
\textbf{Representative keywords and evidence} &
\textbf{Records} \\
\hline
Tropical Cyclone &
hurricane, typhoon, tropical storm, cyclone, Harvey, Irma, Maria, Ida, Ian, Laura, Sally, Fiona, Idalia &
8,676 \\

Flood &
flood, flooding, flooded, inundation, high water, flash flood &
1,581 \\

Tornado &
tornado, twister, funnel cloud, tornado damage &
1,071 \\

Extreme Weather &
severe weather, storm damage, wind damage, hail, thunderstorm, strong wind &
785 \\

Earthquake &
earthquake, seismic, quake, Puerto Rico earthquake &
350 \\

Wildfire &
wildfire, fire, forest fire, burn, burned area, smoke &
56 \\

Other or Unknown &
insufficient or unclear textual evidence for a specific disaster type &
144 \\
\hline
\textbf{Total} & & \textbf{12,663} \\
\hline
\end{tabular}
\end{table}

To characterize the disaster context of the archive, records were classified from available textual metadata, including incident names, image paths, folder names, titles, descriptions, and related attributes. Using GPT-assisted semantic screening assigned records to seven categories, as listed in Table \ref{tab:disaster_category_annotation}, \textbf{tropical cyclone, flood, tornado, earthquake, severe weather, fire and wildfire}, and \textbf{other and unknown}. Specifically, tropical cyclones encompass hurricanes, typhoons, and tropical storms, while tornadoes were separated from broader severe-weather events. Where multiple hazards were indicated, the most explicit incident description determined the category. However, records without sufficient evidence were conservatively assigned to other and unknown to avoid confusion in the following longitudinal analysis. In this context, the resulting taxonomy supports characterization of the longitudinal archive and stratified evaluation across different disaster types.

To facilitate reproducibility, we contribute the curated dataset to the community, where all data and code are made openly available at  \url{https://github.com/bobleegogogo/PhotoMappers-VLM}.

\section{Method}

\subsection{Cross-view Geolocalization of Disaster Photos}

As a first step, we employ a retrieval-based cross-view geolocalization approach on the PhotoMappers archive. Each geolocation contains co-located multi-view observations, including ground-level disaster imagery (VGI), SVI, and RSI.

Given a VGI query, the objective is to recover its geolocation by exploiting complementary visual evidence across these views. Specifically, cross-view geolocalization retrieves candidate locations from a georeferenced reference gallery, where the coordinates associated with the RSI provide the geographic reference for each candidate. We follow a similar architecture to \citet{yin2026triple}, including a two-stage VGI$\rightarrow$SVI and SVI$\rightarrow$RSI retrieval setting and multi-view configurations that jointly use VGI, SVI, and RSI.

We adopt DINOv3~\citep{simeoni2025dinov3} as the visual backbone for cross-view imagery encoding. Each image (VGI, SVI, or RSI) is encoded into a shared embedding space, and the similarity between a query embedding $L_c$ and a candidate reference embedding $L_r^i$ is measured using cosine similarity:

%----------------------------------------------
\begin{equation}
\label{eq:cosine_similarity}
S(L_c,L_r^i)=
\frac{L_c \cdot L_r^i}
{\|L_c\|\|L_r^i\|}.
\end{equation}
%----------------------------------------------

Based on the similarity scores, candidate references are ranked from the most to the least similar. Geolocalization performance is evaluated using Recall@K (i.e., R@K), which measures the percentage of VGI queries for which the correct georeferenced location appears among the top $K$ retrieved candidates. Specifically, R@1 indicates how often the correct location is ranked first, while R@5 and R@10 indicate how often it appears within the top five and top ten candidates, respectively. We additionally report R@1\%, which measures whether the correct location occurs within the top 1\% of the ranked reference gallery and therefore provides a retrieval measure relative to gallery size. Higher values indicate better geolocalization performance.

Finally, we apply Sinkhorn-based re-ranking~\citep{sinkhorn1967concerning,cuturi2013sinkhorn} as a post-processing step under the one-to-one matching setting. Independent ranking may produce hub references that receive high similarity scores from multiple queries, then Sinkhorn re-ranking introduces a global matching constraint to reduce such assignments and refine the final retrieval ranking.

\subsection{Geolocalization Reasoning Indicators Design}

For geospatial reasoning of disaster photos, we design an indicator-based reasoning framework to evaluate the quality and validity of VLM-generated explanations for VGI--SVI--RSI cross-view geolocalization results. Herein, these 30 indicators were developed through empirical visual interpretation of disaster images across different disaster types and scene categories. As a results, our domain experts discussed and organized recurring visual characteristics and spatial relationships into three dimensions: 1). Visual Appearance, 2). Spatial Layout, and 3). Distinctive Cue (listed in Table~\ref{tab:vgi_svi_rsi_indicators_long}). Then, repeated reviews were conducted to identify missing cues and improve indicator coverage. These 30 indicators are further classified into five semantic classes, including building and structure, road and transport, vegetation and terrain, objects and facilities, and global scene. The resulting 30 indicators provide an initial, extensible set of stable visual and spatial cues. They represent cues commonly used in human cross-view interpretation and support the quality assessment of spatial consistency between a VGI image and its candidate reference pair.

%----------------------------------------------
\begin{table*}[!ht]
\centering
\caption{Indicators for evaluating VLM-based cross-view geographic reasoning.}
\label{tab:vgi_svi_rsi_indicators_long}

\scriptsize
\renewcommand{\arraystretch}{1.05}
\setlength{\tabcolsep}{2pt}

\begin{adjustbox}{max width=0.98\textwidth,center}
\begin{tabular}{
>{\centering\arraybackslash}p{0.030\textwidth}
>{\raggedright\arraybackslash}p{0.31\textwidth}
>{\raggedright\arraybackslash}p{0.31\textwidth}
>{\centering\arraybackslash}p{0.085\textwidth}
>{\centering\arraybackslash}p{0.085\textwidth}
>{\centering\arraybackslash}p{0.085\textwidth}
}

\hline
\multirow[c]{2}{*}{\textbf{No.}} &
\multirow[c]{2}{*}{\textbf{Indicator}} &
\multirow[c]{2}{*}{\textbf{Geographic interpretation}} &
\multicolumn{3}{c}{\textbf{Reasoning dimension}} \\
\cline{4-6}
& & &
\scriptsize\textbf{Visual} &
\scriptsize\textbf{Spatial} &
\scriptsize\textbf{Distinctive} \\
\hline

1 & Building type and scale
& Similarity in building form and scale
& $\checkmark$ & & \\

2 & Overall building appearance
& Similarity in architectural appearance
& $\checkmark$ & & \\

3 & Building position relative to nearby roads or open space
& Relative location within the local built environment
& & $\checkmark$ & \\

4 & Building position and shape correspondence in SVI--RSI
& Cross-view correspondence in building geometry
& & $\checkmark$ & \\

5 & Recognizable building outline
& Distinctive building morphology
& & & $\checkmark$ \\

6 & Distinctive building appearance or visible name
& Visually identifiable building characteristics
& & & $\checkmark$ \\

\hline

7 & Road type and width
& Similarity in road form and hierarchy
& $\checkmark$ & & \\

8 & Road markings and boundary features
& Similarity in road-surface and edge characteristics
& $\checkmark$ & & \\

9 & Road direction and connectivity
& Road orientation and network configuration
& & $\checkmark$ & \\

10 & Road position relative to surrounding elements
& Spatial relation between roads and nearby features
& & $\checkmark$ & \\

11 & Distinctive intersection, bridge, or road curve
& Recognizable transport-network geometry
& & & $\checkmark$ \\

12 & Traffic lights, road signs, or road markers
& Locally distinctive transport features
& & & $\checkmark$ \\

\hline

13 & Vegetation coverage and type
& Similarity in vegetation and land-cover characteristics
& $\checkmark$ & & \\

14 & Ground surface characteristics
& Similarity in terrain and surface conditions
& $\checkmark$ & & \\

15 & Vegetation position relative to roads or buildings
& Spatial relation between vegetation and built features
& & $\checkmark$ & \\

16 & Waterbody, shoreline, or open-space position
& Spatial configuration of natural and open-space features
& & $\checkmark$ & \\

17 & Prominent trees, forest belts, or distinctive vegetation
& Recognizable vegetation patterns
& & & $\checkmark$ \\

18 & Coastline, river, lake, or slope outline
& Distinctive natural morphology
& & & $\checkmark$ \\

\hline

19 & Utility poles, streetlights, and cables
& Similarity in persistent infrastructure
& $\checkmark$ & & \\

20 & Road signs, fences, and guardrails
& Similarity in streetscape objects
& $\checkmark$ & & \\

21 & Facility position relative to roads or buildings
& Spatial relation between facilities and built features
& & $\checkmark$ & \\

22 & Local facility grouping and arrangement
& Spatial organization of infrastructure
& & $\checkmark$ & \\

23 & Large facilities such as antennas, water towers, or pylons
& Distinctive infrastructure landmarks
& & & $\checkmark$ \\

24 & Recognizable signs such as storefront signs, billboards, or gas-station signs
& Place-specific visual identifiers
& & & $\checkmark$ \\

\hline

25 & Scene functional type
& Similarity in land-use and scene function
& $\checkmark$ & & \\

26 & Spatial density and openness
& Similarity in urban density and openness
& $\checkmark$ & & \\

27 & Overall building--road--vegetation relationship
& Configuration of major landscape elements
& & $\checkmark$ & \\

28 & Consistency between the VGI visible scene and the SVI--RSI overall scene
& Overall cross-view spatial correspondence
& & $\checkmark$ & \\

29 & Large building complex, community, or commercial area
& Distinctive built-environment context
& & & $\checkmark$ \\

30 & Coastline, river, bridge, or regional background
& Distinctive regional geographic context
& & & $\checkmark$ \\

\hline
\end{tabular}
\end{adjustbox}

\end{table*}
%----------------------------------------------

\subsection{VLM-based Geospatial Reasoning and Evaluation}

Based on the aforementioned empirical indicator framework, geospatial reasoning quality is evaluated using a criterion-based scoring scheme. For each indicator, the VLM-generated reasoning is assigned a score of correct (2), partially correct (1), or incorrect (0), based on whether the indicator provides consistent visual evidence of a match between the VGI image and the corresponding SVI-RSI reference pair. A score of 2 indicates that the reasoning is directly supported by visual evidence from the VGI image and at least one reference view, without clear contradictions from other views. A score of 1 represents reasoning that is partially supported but affected by incomplete, weak, or ambiguous cross-view evidence. A score of 0 indicates that the relevant feature is not visible or matched across the compared views, or that the reasoning is unsupported, contradicted by visual evidence, spatially inconsistent, or hallucinated.

All VLM-based annotation and evaluation are conducted in a zero-shot setting without training on disaster-specific examples. Qwen-3.8-Max \citep{qwen2026qwen38} is used as the annotation and VLM-as-a-Judge model \citep{wu2022survey,lin2026act}. As the latest VLM, Qwen-3.8-Max ranks among the top models on the Vision Arena Leaderboard \citep{arena2026vision} and provides a good balance between performance and cost. Its annotations are used as ground truth after a careful human-in-the-loop expert verification and examination. For human-in-the-loop validation, 10\% of the samples are stratified according to predicted damage severity levels and independently reviewed by two human experts. The experts examine whether the predicted spatial matching decisions and corresponding reasoning are visually grounded in the VGI, SVI, and RSI evidence. Disagreements are resolved through discussion to identify unsupported explanations, ambiguous cases, and recurring errors. 

%As a context, LLM/VLM-as-a-Judge frameworks with human-in-the-loop verification have been commonly applied in multimodal understanding tasks \citep{wu2022survey,lin2026act}.

Indicator-level scores and overall predictions are evaluated using agreement rate and quadratic weighted kappa (\(\kappa_w\)). Herein, agreement rate measures exact correspondence between VLM-generated and human-verified scores, while \(\kappa_w\) accounts for chance agreement and the ordinal structure of the three-level scoring scheme \(C=\{0,1,2\}\). Compared with conventional Cohen's kappa, quadratic weighting assigns smaller penalties to adjacent-category disagreements and increasingly larger penalties to more substantial deviations \citep{cohen1968weighted}:

\begin{equation}
\kappa_w =
1-\frac{\sum_{i=1}^{K}\sum_{j=1}^{K}w_{ij}O_{ij}}
{\sum_{i=1}^{K}\sum_{j=1}^{K}w_{ij}E_{ij}},
\qquad
w_{ij}=\frac{(i-j)^2}{(K-1)^2},
\end{equation}

where \(O_{ij}\) and \(E_{ij}\) denote the observed and expected agreement matrices, respectively, and \(K\) is the number of scoring categories. These metrics quantify both exact agreement and the severity of disagreement between VLM predictions and reference annotations.

\subsection{Longitudinal Spatiotemporal Analysis}

To characterize the long-term evolution of PhotoMappers, we conduct a year-wise analysis along three dimensions: \textbf{geographic distribution, thematic composition}, and \textbf{model performance}. First, annual observations are aggregated to quantify changes in data volume and spatial coverage. For the contiguous United States, the collected 12,663 PhotoMappers records are spatially joined to state boundaries. Year-wise choropleth maps are produced using a common spatial extent, classification scheme, and color scale, allowing direct comparison of the geographic concentration and expansion or contraction of crowdsourced disaster mapping over states and years.

Second, temporal changes in the composition of PhotoMappers archieve are further examined by disaster types and FEMA Community Lifelines. For each category \(c\) in year \(t\), the annual proportion is calculated as
\begin{equation}
P_{c,t}=\frac{N_{c,t}}{N_t},
\end{equation}
where \(N_{c,t}\) denotes the number of records belongs to category \(c\) in year \(t\), and \(N_t\) is the total number of valid record in that year. These proportions are represented using 100\% stacked bar charts, thereby separating compositional change from variation in annual sample size.

Finally, we examine whether the performance of GeoAI and VLM varies over time and across the changing composition of the archive. Annual performance curves summarize cross-view geolocalization accuracy and VLM-based geospatial reasoning agreement, followed by longitudinal analyses across disaster types and geospatial reasoning evidences.  Herein, geospatial reasoning evidences include 1) appearance, 2) spatial layout, and 3) landmark cues. This longitudinal analysis jointly evaluates changes in where disaster observations were produced, what disaster and community impacts they represented, and how consistently GeoAI and VLM together reproduced human geographic interpretation across years and contexts.

%================================================================
\section{Experimental Results}
\label{sec:results}
%================================================================

\subsection{Cross-view Geolocalization Results}
\label{sec:geolocation_results}

Table~\ref{tab:cross_view_geolocation} reports cross-view geolocalization
performance on the test split, including  2,347 samples. It is obvious that the tri-view (i.e., VGI--SVI--RSI) configuration outperforms both two-view alternatives, achieving 36.17\% R@1, 57.82\% R@5, and 65.83\% R@10, compared with 27.82\%, 47.59\%, and 57.95\% for VGI--RSI and 25.56\%, 45.85\%, and 54.67\% for VGI--SVI. Moreover, adding Sinkhorn-based re-ranking further improves performance to 40.90\% R@1, 61.10\% R@5, and 69.45\% R@10. These results indicate that VGI, SVI, and RSI provide complementary geographic evidence, while post-process re-ranking further improves cross-view retrieval accuracy.

%--------------------------------------------------
\begin{table*}[!ht]
\centering
\caption{Cross-view geolocalization performance on the test
split. Results are reported as recall (\%). Herein, R@1\% denotes the percentage of
queries for which the correct location is retrieved within the top 1\% of
the candidate set. Bold values indicate the best accuracy.}
\label{tab:cross_view_geolocation}

\small
\renewcommand{\arraystretch}{1.0}
\setlength{\tabcolsep}{12pt}

\begin{tabular}{lcccc}
\toprule
\textbf{Cross-View Models}
& \textbf{R@1}
& \textbf{R@5}
& \textbf{R@10}
& \textbf{R@1\%} \\
\midrule

VGI--RSI
& 27.82
& 47.59
& 57.95
& 68.51 \\

VGI--SVI
& 25.56
& 45.85
& 54.67
& 64.89 \\

VGI--SVI--RSI
& 36.17
& 57.82
& 65.83
& 74.52 \\

VGI--SVI--RSI + Sinkhorn
& \textbf{40.90}
& \textbf{61.10}
& \textbf{69.45}
& \textbf{77.46} \\

\bottomrule
\end{tabular}
\end{table*}
%--------------------------------------------------

%================================================================
\subsection{VLM-based Geospatial Reasoning Results}
\label{sec:reasoning_results}

Next, we evaluate VLM-based geographic reasoning using the aforementioned 30-indicator framework, organized by 5 Geographic Cues and 3 Reasoning Evidences dimensions (as shown in Table~\ref{tab:vlm_indicator_summary}). As a result, Qwen-3.8-Max shows high overall agreement with both human experts: $\kappa_w=0.97$ with Expert A and 0.94 with Expert B, compared with an inter-expert agreement of 0.95. The corresponding agreement rates are 95.76\%, 96.29\%, and 93.43\%, respectively. As a result, this comparable VLM-expert and inter-expert agreement supports the use of the validated Qwen-3.8-Max annotations as a scalable reference for subsequent reasoning evaluation.

Based on this reference, Qwen3.6-27B achieves an overall $\kappa_w=0.65$ and 67.20\% agreement, compared with 0.30 and 51.04\% for InternVL3-8B. Noticeably, performance also varies significantly across different Geographic Cues and Reasoning Evidences (as shown in Table~\ref{tab:vlm_indicator_summary}). For Qwen3.6-27B, Spatial Layout reaches $\kappa_w=0.75$, compared with 0.52 for Landmark Cue; the corresponding values for InternVL3-8B are 0.36 and 0.23. At the indicator level, as shown in Table \ref{tab:vlm_indicator_agreement}, VGI-SVI-RSI scene consistency reaches $\kappa_w=0.85$ for Qwen3.6-27B, whereas several facility-, sign-, and landmark-based indicators show substantially lower agreement. This phenomena motivates us to extend this into a longitudinal analysis to examine when and where AI struggles for accurate geospatial reasoning.

% Required packages: booktabs, multirow, array, graphicx
%————————————————————————————————————————————————————————————
\begin{table*}[!htbp]
\centering
\caption{Validation of VLM-based geospatial reasoning regarding geographic cue and evidence. Herein, VLM in the table refers to Qwen3.8-Max, with expert-reviewed samples used as ground truth.}
\label{tab:vlm_indicator_summary}
\footnotesize
\renewcommand{\arraystretch}{1.00}
\setlength{\tabcolsep}{3pt}

\resizebox{\textwidth}{!}{%
\begin{tabular}{
>{\centering\arraybackslash}p{0.33cm}
>{\raggedright\arraybackslash}p{4.5cm}
!{\vrule width 0.6pt}
cc cc cc
!{\vrule width 0.6pt}
cc cc
}
\toprule

\multirow{2}{*}{\textbf{No.}}
& \multirow{2}{*}{\textbf{Indicator name}}
& \multicolumn{2}{c}{\textbf{VLM vs Exp. A}}
& \multicolumn{2}{c}{\textbf{VLM vs Exp. B}}
& \multicolumn{2}{c!{\vrule width 0.6pt}}{\textbf{Exp. A vs Exp. B}}
& \multicolumn{2}{c}{\shortstack{\textbf{Qwen3.6-27B}}}
& \multicolumn{2}{c}{\shortstack{\textbf{InternVL3-8B}}} \\

\cmidrule(lr){3-4}
\cmidrule(lr){5-6}
\cmidrule(lr){7-8}
\cmidrule(lr){9-10}
\cmidrule(lr){11-12}

&
& \textbf{$\kappa_w$} & \textbf{Agr.}
& \textbf{$\kappa_w$} & \textbf{Agr.}
& \textbf{$\kappa_w$} & \textbf{Agr.}
& \textbf{$\kappa_w$} & \textbf{Agr.}
& \textbf{$\kappa_w$} & \textbf{Agr.} \\
\midrule

\multicolumn{12}{l}{\textbf{Geographic Cue}} \\

\multicolumn{2}{l!{\vrule width 0.6pt}}{Building and structure average}
& 0.97 & 96.01\%
& 0.93 & 96.48\%
& 0.96 & 94.13\%
& 0.74 & 72.49\%
& 0.28 & 49.21\% \\

\multicolumn{2}{l!{\vrule width 0.6pt}}{Road and transport average}
& 0.97 & 95.85\%
& 0.95 & 96.95\%
& 0.96 & 94.20\%
& 0.69 & 67.80\%
& 0.28 & 46.27\% \\

\multicolumn{2}{l!{\vrule width 0.6pt}}{Vegetation and terrain average}
& 0.97 & 95.46\%
& 0.95 & 96.64\%
& 0.95 & 93.27\%
& 0.60 & 62.03\%
& 0.27 & 48.84\% \\

\multicolumn{2}{l!{\vrule width 0.6pt}}{Objects and facilities average}
& 0.97 & 96.17\%
& 0.94 & 95.93\%
& 0.95 & 93.58\%
& 0.53 & 60.63\%
& 0.26 & 49.54\% \\

\multicolumn{2}{l!{\vrule width 0.6pt}}{Global scene average}
& 0.97 & 95.31\%
& 0.95 & 95.46\%
& 0.95 & 91.94\%
& 0.69 & 73.06\%
& 0.41 & 61.36\% \\

\midrule

\multicolumn{12}{l}{\textbf{Reasoning Evidence}} \\

\multicolumn{2}{l!{\vrule width 0.6pt}}{Appearance dimension average}
& 0.96 & 95.35\%
& 0.95 & 96.53\%
& 0.95 & 93.00\%
& 0.67 & 66.31\%
& 0.31 & 50.74\% \\

\multicolumn{2}{l!{\vrule width 0.6pt}}{Spatial layout dimension average}
& 0.97 & 95.63\%
& 0.94 & 95.96\%
& 0.96 & 93.24\%
& 0.75 & 73.55\%
& 0.36 & 55.18\% \\

\multicolumn{2}{l!{\vrule width 0.6pt}}{Landmark cue dimension average}
& 0.97 & 96.29\%
& 0.94 & 96.38\%
& 0.95 & 94.04\%
& 0.52 & 61.75\%
& 0.23 & 47.21\% \\

\specialrule{0.5pt}{2pt}{2pt}

\multicolumn{2}{l!{\vrule width 0.6pt}}{\textbf{Overall indicator-level average}}
& 0.97 & 95.76\%
& 0.94 & 96.29\%
& 0.95 & 93.43\%
& 0.65 & 67.20\%
& 0.30 & 51.04\% \\

\bottomrule
\end{tabular}%
}

\end{table*}
%————————————————————————————————————————————————————————————

%————————————————————————————————————————————————————————————
\begin{table*}[!htbp]
\centering
\caption{Validation of VLM-based geospatial reasoning regarding 30 indicators. Herein, VLM in the table refers to Qwen3.8-Max, with expert-reviewed samples used as ground truth.}
\label{tab:vlm_indicator_agreement}

\footnotesize
\renewcommand{\arraystretch}{1.00}
\setlength{\tabcolsep}{3pt}

\resizebox{\textwidth}{!}{%
\begin{tabular}{
>{\centering\arraybackslash}p{0.33cm}
>{\raggedright\arraybackslash}p{4.5cm}
!{\vrule width 0.6pt}
cc cc cc
!{\vrule width 0.6pt}
cc cc
}

\toprule

\multirow{2}{*}{\textbf{No.}}
& \multirow{2}{*}{\textbf{Indicator name}}
& \multicolumn{2}{c}{\textbf{VLM vs Exp. A}}
& \multicolumn{2}{c}{\textbf{VLM vs Exp. B}}
& \multicolumn{2}{c!{\vrule width 0.6pt}}{\textbf{Exp. A vs Exp. B}}
& \multicolumn{2}{c}{\shortstack{\textbf{Qwen3.6-27B}}}
& \multicolumn{2}{c}{\shortstack{\textbf{InternVL3-8B}}} \\

\cmidrule(lr){3-4}
\cmidrule(lr){5-6}
\cmidrule(lr){7-8}
\cmidrule(lr){9-10}
\cmidrule(lr){11-12}

&
& \textbf{$\kappa_w$} & \textbf{Agr.}
& \textbf{$\kappa_w$} & \textbf{Agr.}
& \textbf{$\kappa_w$} & \textbf{Agr.}
& \textbf{$\kappa_w$} & \textbf{Agr.}
& \textbf{$\kappa_w$} & \textbf{Agr.} \\
\midrule

1 & Building type and scale
& 0.96 & 95.31\%
& 0.95 & 96.24\%
& 0.94 & 92.49\%
& 0.72 & 68.65\%
& 0.32 & 52.67\% \\

2 & Overall building appearance
& 0.98 & 96.71\%
& 0.94 & 96.24\%
& 0.96 & 94.84\%
& 0.78 & 75.45\%
& 0.30 & 51.27\% \\

3 & Building-road-open space position
& 0.97 & 95.77\%
& 0.94 & 95.77\%
& 0.96 & 93.43\%
& 0.75 & 72.63\%
& 0.34 & 55.20\% \\

4 & Building shape correspondence
& 0.95 & 93.90\%
& 0.93 & 96.71\%
& 0.94 & 92.49\%
& 0.77 & 75.96\%
& 0.30 & 50.28\% \\

5 & Recognizable building outline
& 0.98 & 96.71\%
& 0.93 & 96.24\%
& 0.96 & 94.84\%
& 0.73 & 74.17\%
& 0.26 & 47.87\% \\

6 & Distinctive building appearance
& 0.98 & 97.64\%
& 0.92 & 97.64\%
& 0.97 & 96.71\%
& 0.65 & 68.11\%
& 0.18 & 37.99\% \\

\midrule

7 & Road type and width
& 0.97 & 95.77\%
& 0.97 & 97.65\%
& 0.96 & 94.37\%
& 0.73 & 68.84\%
& 0.33 & 53.09\% \\

8 & Road markings and boundaries
& 0.97 & 96.24\%
& 0.96 & 98.12\%
& 0.96 & 95.31\%
& 0.69 & 64.76\%
& 0.28 & 43.39\% \\

9 & Road direction and connectivity
& 0.96 & 93.87\%
& 0.94 & 96.23\%
& 0.95 & 91.98\%
& 0.79 & 74.96\%
& 0.33 & 51.20\% \\

10 & {\fontsize{7.5pt}{9pt}\selectfont Road-element position relationship}
& 0.97 & 95.77\%
& 0.94 & 95.77\%
& 0.96 & 93.43\%
& 0.79 & 77.88\%
& 0.33 & 56.19\% \\

11 & Distinctive intersections or curves
& 0.98 & 97.18\%
& 0.92 & 96.24\%
& 0.96 & 95.31\%
& 0.66 & 68.13\%
& 0.20 & 38.61\% \\

12 & Traffic signs and markers
& 0.97 & 96.24\%
& 0.95 & 97.65\%
& 0.95 & 94.84\%
& 0.48 & 52.23\%
& 0.21 & 35.11\% \\

\midrule

13 & Vegetation coverage and type
& 0.96 & 95.77\%
& 0.95 & 96.24\%
& 0.94 & 92.96\%
& 0.65 & 65.51\%
& 0.34 & 53.61\% \\

14 & Ground surface characteristics
& 0.95 & 92.96\%
& 0.95 & 96.71\%
& 0.94 & 91.08\%
& 0.66 & 62.09\%
& 0.32 & 47.66\% \\

15 & Vegetation-road-building position
& 0.97 & 95.77\%
& 0.95 & 96.24\%
& 0.96 & 93.43\%
& 0.74 & 71.65\%
& 0.34 & 54.78\% \\

16 & {\fontsize{7.5pt}{9pt}\selectfont Waterbody and open-space position}
& 0.97 & 96.24\%
& 0.95 & 96.71\%
& 0.95 & 93.90\%
& 0.56 & 54.45\%
& 0.25 & 42.22\% \\

17 & Prominent vegetation features
& 0.98 & 96.71\%
& 0.96 & 96.71\%
& 0.96 & 94.37\%
& 0.66 & 63.64\%
& 0.24 & 45.50\% \\

18 & Coastline and terrain outline
& 0.96 & 95.31\%
& 0.97 & 97.18\%
& 0.95 & 93.90\%
& 0.34 & 54.83\%
& 0.12 & 49.30\% \\

\midrule

19 & Utility poles and cables
& 0.96 & 94.84\%
& 0.94 & 95.31\%
& 0.93 & 91.08\%
& 0.59 & 60.50\%
& 0.33 & 49.48\% \\

20 & Road signs, fences, and guardrails
& 0.97 & 95.77\%
& 0.95 & 98.12\%
& 0.96 & 95.31\%
& 0.53 & 57.87\%
& 0.23 & 41.19\% \\

21 & Facility-road-building position
& 0.98 & 97.18\%
& 0.93 & 95.31\%
& 0.96 & 94.37\%
& 0.73 & 70.20\%
& 0.36 & 53.33\% \\

22 & Facility grouping arrangement
& 0.98 & 97.18\%
& 0.95 & 96.24\%
& 0.96 & 94.84\%
& 0.71 & 69.07\%
& 0.29 & 47.33\% \\

23 & Large facility structures
& 0.97 & 96.71\%
& 0.92 & 94.84\%
& 0.93 & 92.96\%
& 0.21 & 49.67\%
& 0.15 & 57.69\% \\

24 & Recognizable commercial signs
& 0.97 & 95.31\%
& 0.93 & 95.77\%
& 0.95 & 92.96\%
& 0.39 & 56.45\%
& 0.20 & 48.20\% \\

\midrule

25 & Scene functional type
& 0.96 & 94.84\%
& 0.95 & 95.31\%
& 0.94 & 91.08\%
& 0.70 & 71.74\%
& 0.37 & 60.17\% \\

26 & Spatial density and openness
& 0.96 & 95.31\%
& 0.94 & 95.31\%
& 0.93 & 91.55\%
& 0.64 & 67.67\%
& 0.32 & 54.83\% \\

27 & {\fontsize{7.5pt}{9pt}\selectfont Building-road-vegetation relation}
& 0.97 & 95.31\%
& 0.94 & 94.84\%
& 0.95 & 92.02\%
& 0.79 & 80.55\%
& 0.47 & 66.45\% \\

28 & VGI-SVI-RSI scene consistency
& 0.97 & 95.31\%
& 0.95 & 95.77\%
& 0.96 & 92.49\%
& 0.85 & 88.14\%
& 0.58 & 74.84\% \\

29 & Large building complexes
& 0.97 & 95.31\%
& 0.94 & 94.84\%
& 0.94 & 91.55\%
& 0.63 & 72.12\%
& 0.38 & 60.12\% \\

30 & Coastline and regional background
& 0.97 & 95.77\%
& 0.97 & 96.71\%
& 0.95 & 92.96\%
& 0.50 & 58.15\%
& 0.35 & 51.73\% \\

\bottomrule
\end{tabular}%
}

\end{table*}
%————————————————————————————————————————————————————————————

\newpage

An important take-away from the VLM-based reasoning results is that a consistent gap between reasoning over broad spatial configurations and reasoning from localized and place-specific geographic evidence, highlighting the fact that VLMs are still largely biased and insufficient in understanding certain aspects (e.g., object and facilities or landmark cue) of disaster photos. Once more, this finding highlight a key implication that volunteer's effort remains unique and difficult to replace in those indicators and dimensions where VLMs still struggles.

%================================================================
\subsection{Insights from Longitudinal Analysis}
\label{sec:longitudinal_results}

\subsubsection{Evolving Geographies of Disaster Mapping}

\begin{figure}[!ht]
    \centering
    \includegraphics[width=0.6\linewidth]{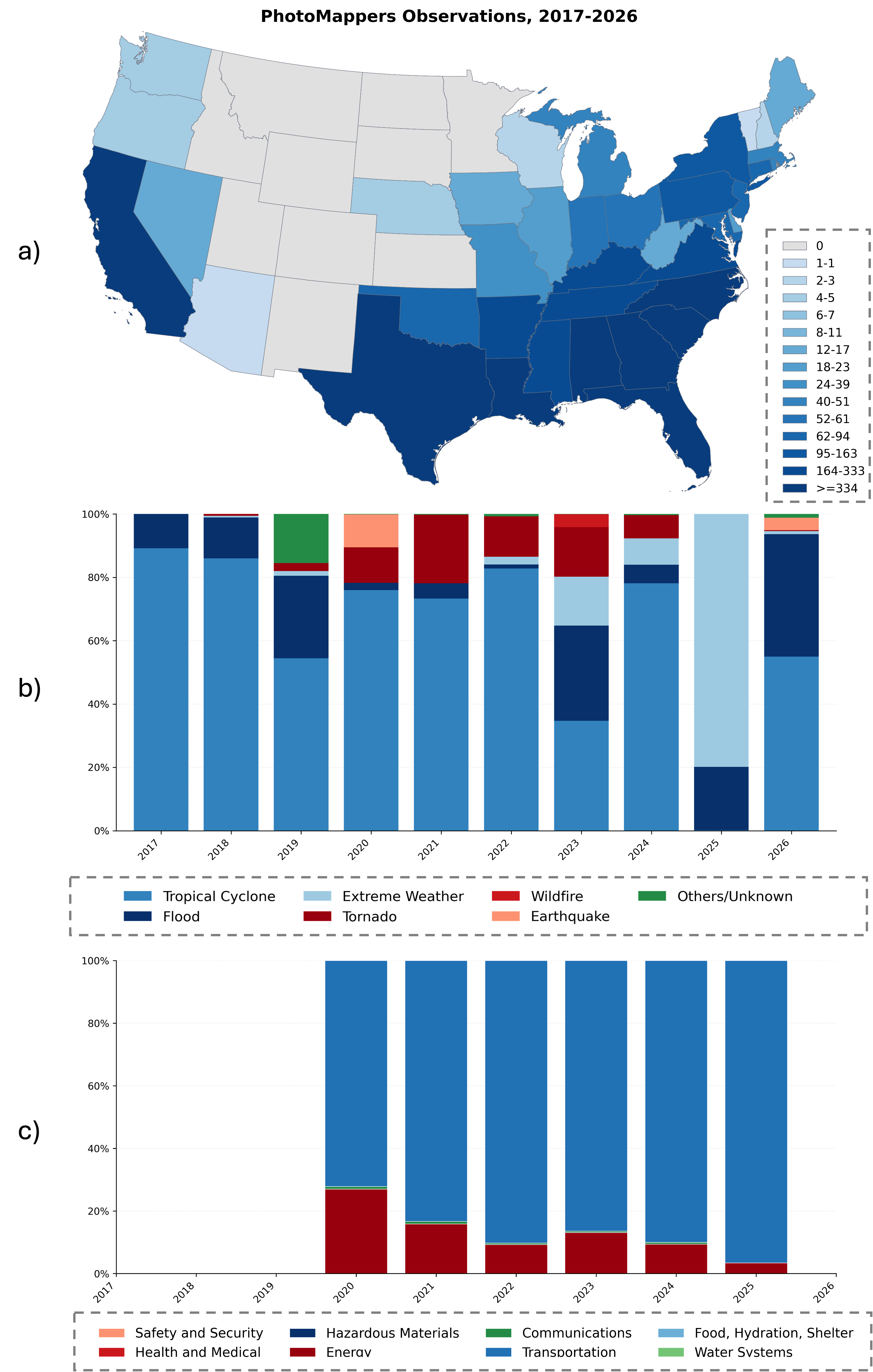}
    \caption{Spatial and temporal distributions of PhotoMappers observations: (a) state-level distribution; (b) annual disaster-type composition; and (c) annual community-lifeline composition.}
    \label{fig:longitudinal_total_distribution}
\end{figure}

\begin{figure}[!ht]
    \centering
    \includegraphics[width=0.8\linewidth]{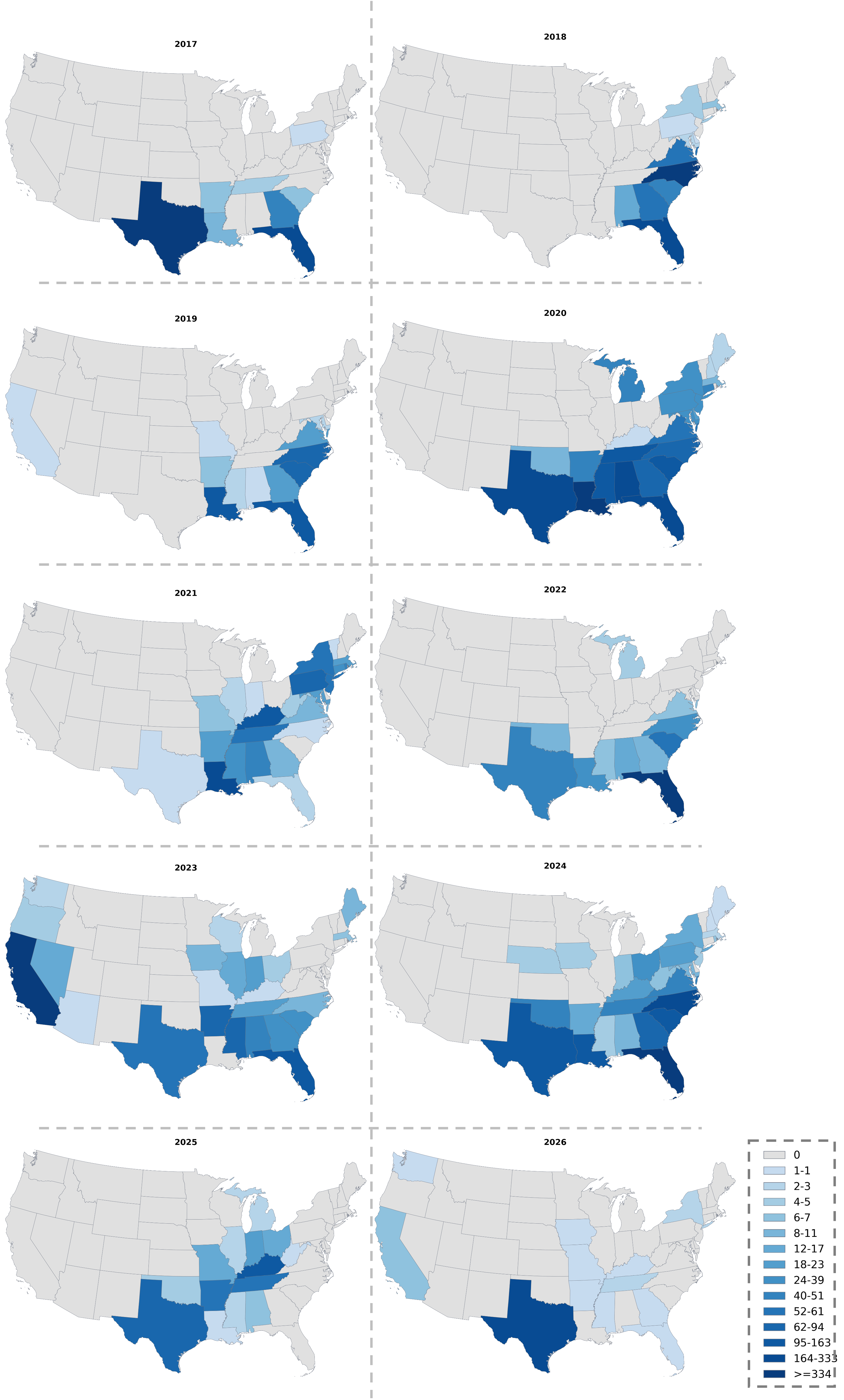}
    \caption{Year-wise geographic distribution of PhotoMappers observations across the contiguous United States. State-level choropleth maps show changes in the spatial footprint and concentration of crowdsourced disaster observations across the study period.}
    \label{fig:longitudinal_yearly_map}
\end{figure}

Splitting the incidents by year, the longitudinal analysis reveals a strongly event-driven geography of PhotoMappers observations. As shown in Figures~\ref{fig:longitudinal_total_distribution} and \ref{fig:longitudinal_yearly_map}, the spatial distribution of PhotoMappers records changes substantially across years, with records concentrated in states and regions affected by specific disaster activations. The decade-long PhotoMappers archive therefore represents the time geography of disasters that were observed, crowdsourced, and mapped rather than a systematic spatial sample of physical disaster occurrence.

%----------------------------------------------
%----------------------------------------------

%----------------------------------------------
%----------------------------------------------

Looking at Figure~\ref{fig:longitudinal_total_distribution}, the thematic composition changes alongside this geographic distributions as well. Annual distributions of disaster types and Community Lifelines show that different years contain diverse combinations of disaster types and infrastructure impacts. These differences also affect the geographic evidence available for localization with some activations are dominated by roads and built environments, whereas others contain stronger signals from vegetation, water, infrastructure, or regional landscape features. Therefore, one can see the longitudinal archive thus captures not only where PhotoMappers operated, but also the changing forms of geographic evidence through which disaster locations were interpreted.

This distinction is important for longitudinal GeoAI and VLM evaluation. Differences in model performance across years cannot be attributed to time alone because each annual subset represents a different combination of hazards, landscapes, and operational priorities. The changing geography of the archive therefore provides the context for interpreting temporal variation in both geolocalization and reasoning performance. We will elaborate on these changing geography in the rest of this section.

%----------------------------------------------------------------
\subsubsection{Uneven Geospatial Reasoning Performance}

\begin{figure}[!ht]
    \centering
    \includegraphics[width=0.6\linewidth]{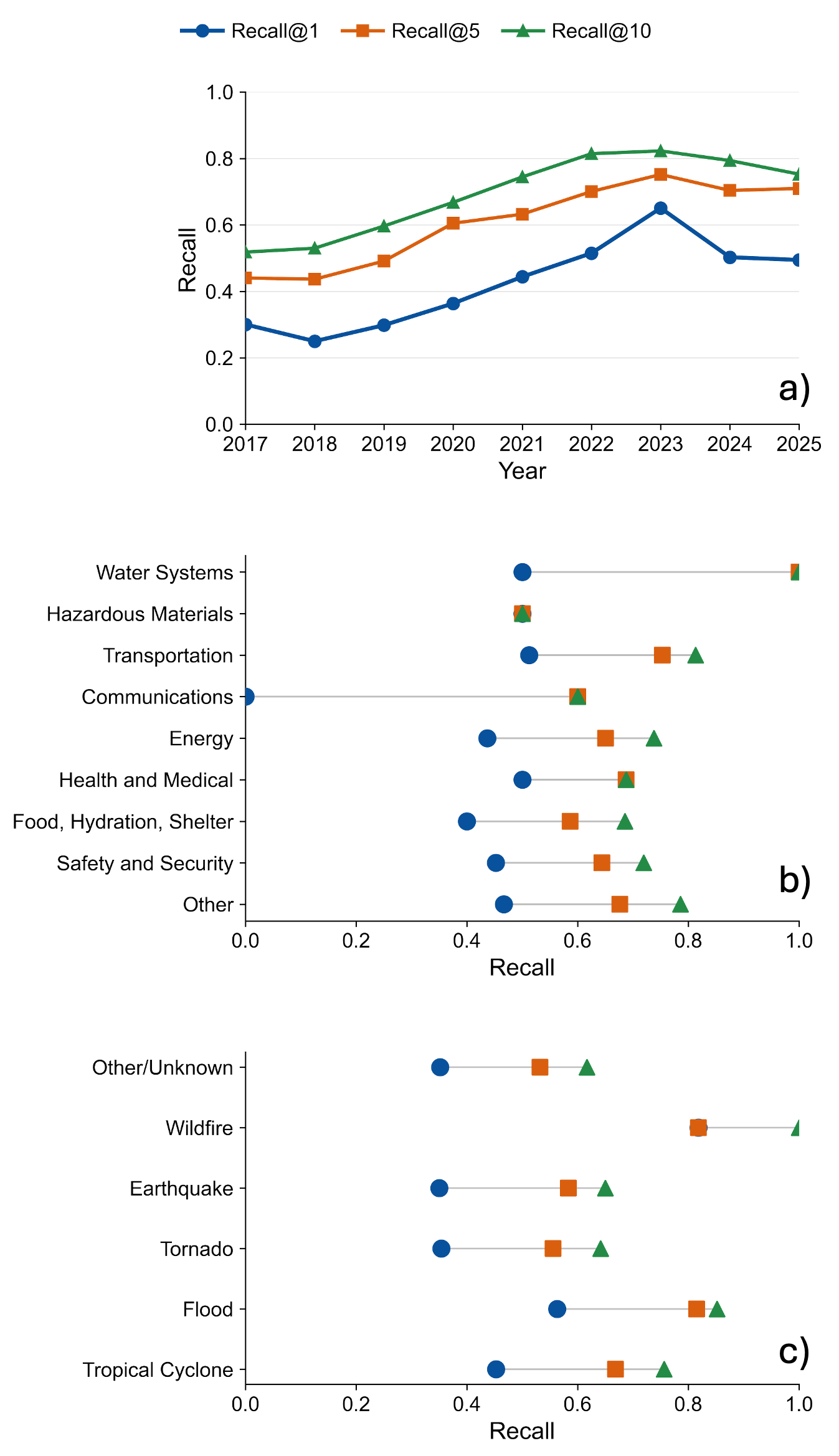}
    \caption{Longitudinal cross-view geolocalization performance from 2017 to 2025. Recall@1, Recall@5, and Recall@10 indicate the proportion of observations for which the correct location is retrieved within the top 1, 5, and 10 candidates, respectively.}
    \label{fig:longitudinal_geolocalization}
\end{figure}

First of all, VLM-based reasoning results varies significantly across the longitudinal period. Figure~\ref{fig:longitudinal_geolocalization} a) shows that R@1 increases from 30.05\% in 2017 to 65.07\% in 2023, before declining to 50.26\% in 2024 and 49.46\% in 2025. R@10 follows a similar trajectory, increasing from 51.81\% in 2017 to 82.27\% in 2023 before decreasing to 75.27\% in 2025. The trajectory is therefore not monotonic, showing later observations are not uniformly easier to geolocate. Instead, retrieval performance varies alongside changes in the geographic and visual composition of the archive.

%----------------------------------------------
%----------------------------------------------

A more persistent pattern is the gap between R@1 and the higher recall levels (also in Figure \ref{fig:longitudinal_geolocalization} a)). For example, in 2022, R@1 is 51.50\%, compared with 70.06\% at R@5 and 81.44\% at R@10. Importantly, this gap shows that the correct location is therefore frequently retrieved within a small candidate set even when it is not ranked first. Geographically, this suggests that the model often recovers a plausible locational context but has greater difficulty discriminating among visually or spatially similar candidate locations.

This phenomena extends across Community Lifelines and disaster types (as shown in Figure \ref{fig:longitudinal_geolocalization} b) and c)). For instance, Transportation reaches 51.25\% R@1 and 81.30\% R@10, whereas Communications increases from 0.00\% R@1 to 60.00\% at R@5 and R@10. Across disaster types, Flood reaches 56.30\% R@1 and 85.19\% R@10, compared with 45.25\% and 75.62\% for Tropical Cyclones. These contrasts show that geolocalization performance varies with the operational and geographic context represented in the PhotoMappers, including the types of infrastructure, landscapes, and disaster impacts available as geographic evidence. 

Herein, results from sparsely represented categories require additional caution. Water Systems and Hazardous Materials contain relatively limited observations, while Wildfire produces a high point estimate of 81.82\% R@1. These values reminds us that the observed samples should not be interpreted as general differences in geolocalization difficulty without accounting for sample size and composition.

Importantly, this longitudinal analysis highlights the real-world complexities that the chaotic and heterogeneous disaster scenes pose to the GeoAI models and general-purpose VLMs, challenging the decision making and deployment process where these gaps are often overlooked. For examples, disaster managers may assume a consistent geolocalization accuracy when planning for their response actions w.r.t different Community Lifeline stabilization (as shown in Figure~\ref{fig:lifelines}), while an unexpected performance drop may results in a significant detour and additional-effort of in-field rescue and construction teams. Herein, the key is to better understand when and where shall we trust volunteers over AI models.

%----------------------------------------------------------------
\subsubsection{Limits of VLM-based Geospatial Reasoning}

\begin{figure}[!ht]
    \centering
    \includegraphics[width=\linewidth]{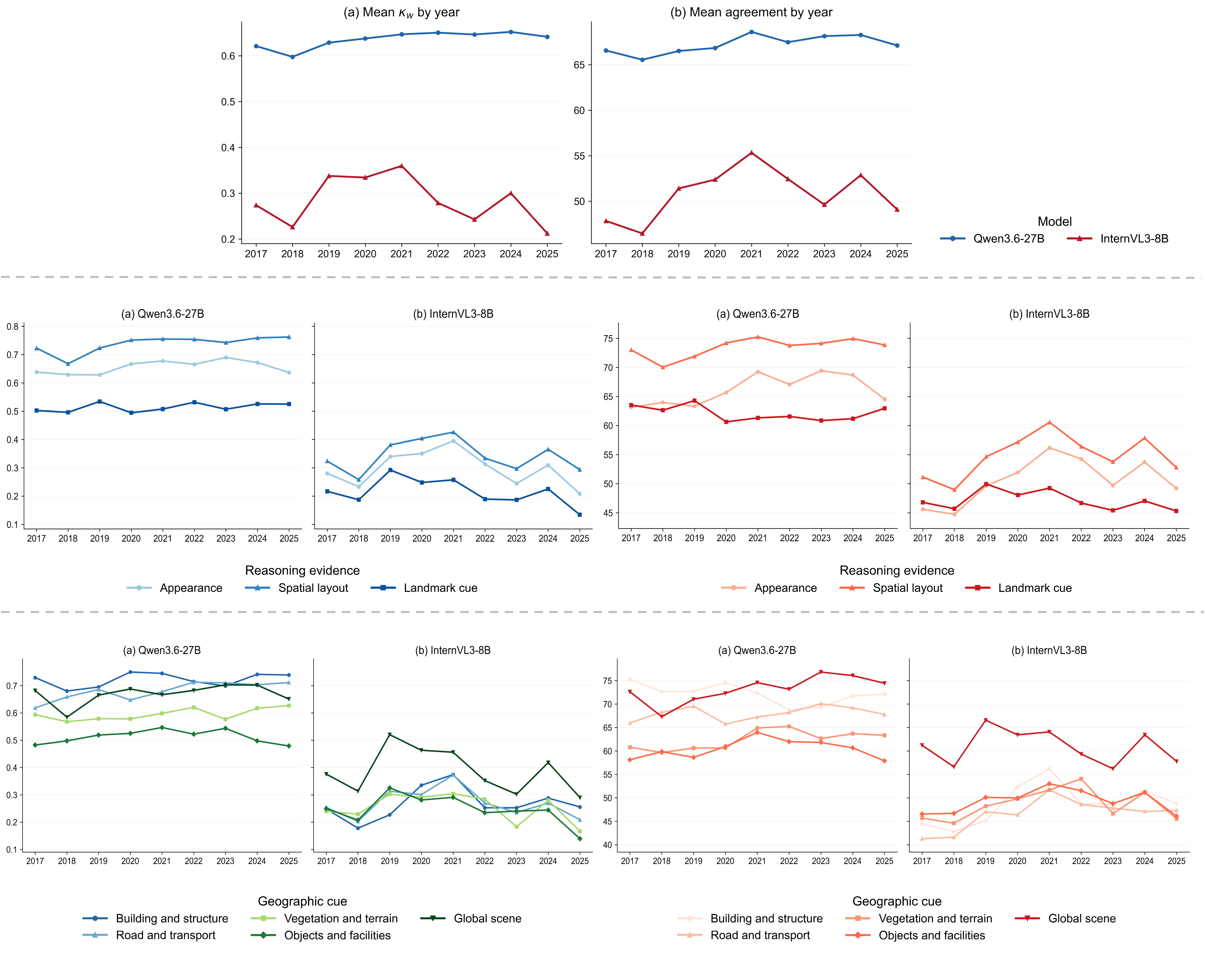}
    \caption{Longitudinal VLM geographic reasoning performance (i.e., a) indicator-wise $\kappa_w$ results and b) indicator-wise agreement) from 2017 to 2025. Year-wise curves summarize agreement across the three Reasoning Evidence dimensions: Visual Appearance, Spatial Layout, and Landmark Cue.}
    \label{fig:longitudinal_yearly_reasoning}
\end{figure}

Following a similar longitudinal analysis, the VLM-based reasoning results reveal a related but distinct geography of model behaviors. Figure~\ref{fig:longitudinal_yearly_reasoning} shows that reasoning agreement varies across years and across forms of geographic evidence. For example, Spatial Layout is more consistently reproduced than Landmark Cue, with Qwen3.6-27B reaches $\kappa_w=0.75$ for Spatial Layout compared with 0.52 for Landmark Cue, while InternVL3-8B reaches 0.36 and 0.23, respectively. 

%----------------------------------------------
%----------------------------------------------

\begin{figure}[!ht]
    \centering
    \includegraphics[width=\linewidth]{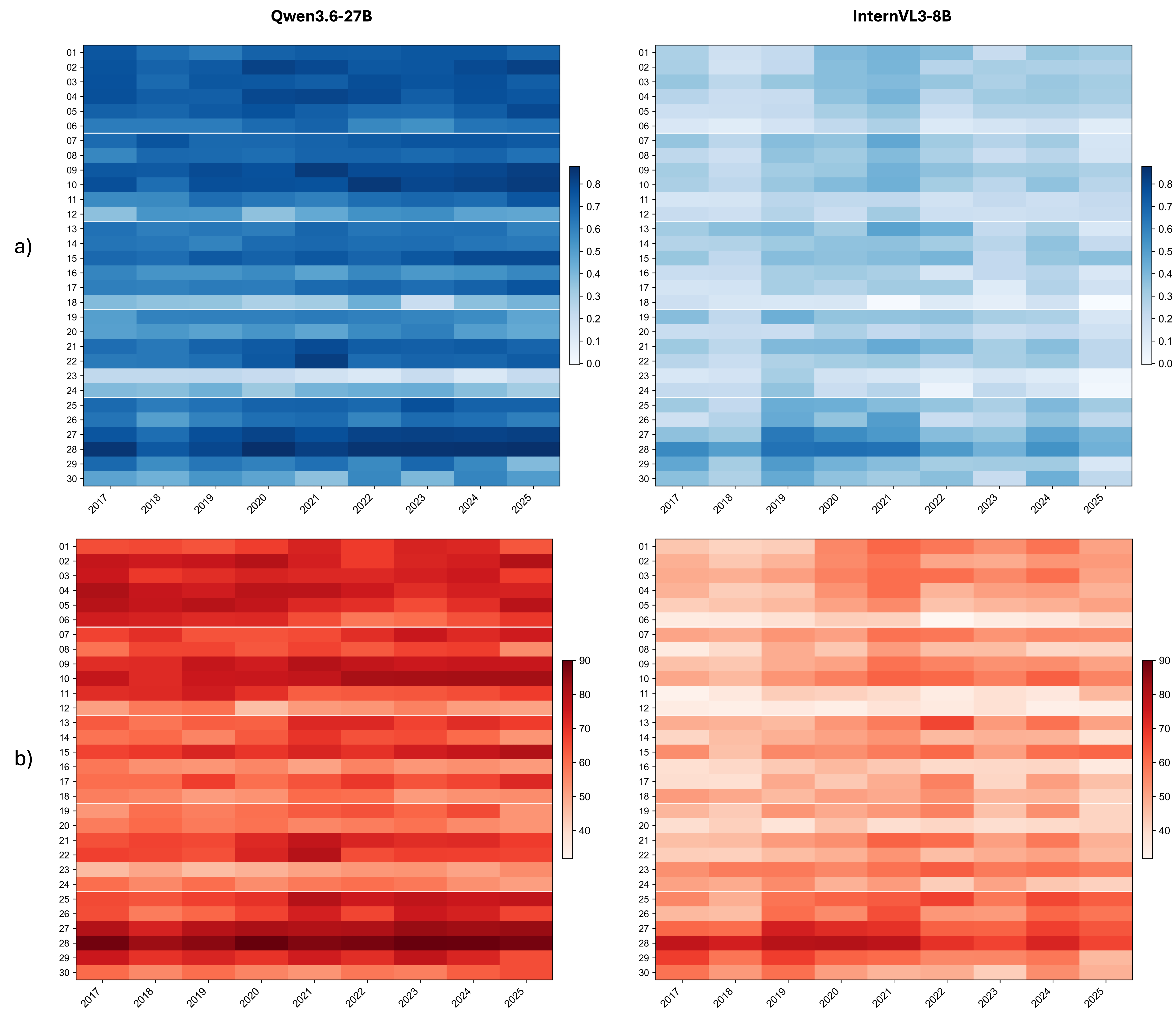}
    \caption{Longitudinal heatmap of indicator-level geographic reasoning performance (i.e., a) indicator-wise $\kappa_w$ results and b) indicator-wise agreement). The 30 indicators are organized by five Geographic Cues: Building and Structure, Road and Transport, Vegetation and Terrain, Objects and Facilities, and Global Scene, and three Reasoning Evidence dimensions: Visual Appearance, Spatial Layout, and Landmark Cue.}
    \label{fig:longitudinal_indicator}
\end{figure}

\begin{figure}[!ht]
    \centering
    \includegraphics[width=\linewidth]{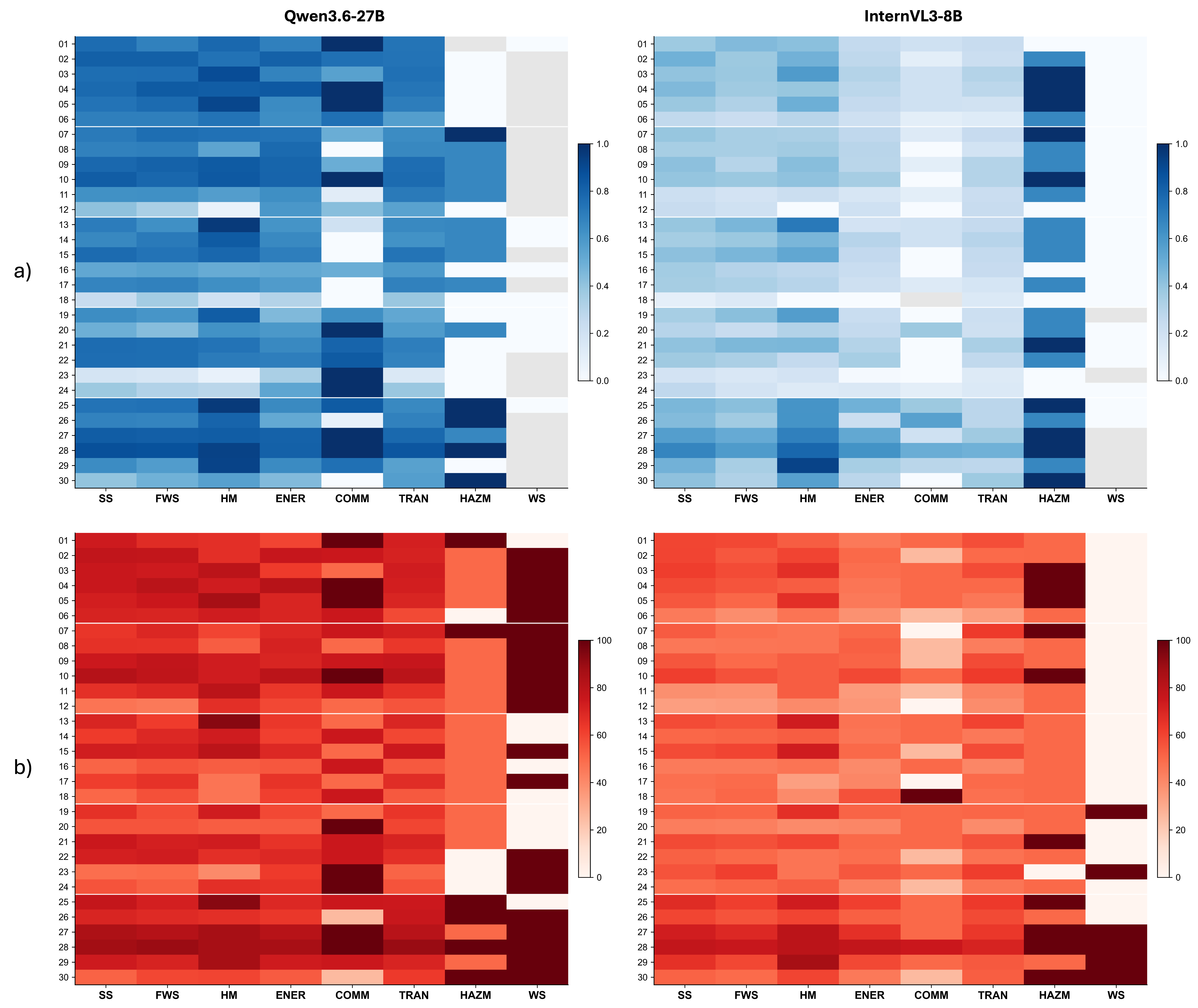}
    \caption{Longitudinal heatmap of indicator-level geographic reasoning performance (i.e., a) indicator-wise $\kappa_w$ and b) indicator-wise agreement) across FEMA Community Lifelines. The comparison shows variation in model performance across different forms of community infrastructure and disaster-response context.}
    \label{fig:longitudinal_lifelines}
\end{figure}

Moreover, the indicator-level heatmaps in Figure~\ref{fig:longitudinal_indicator} and \ref{fig:longitudinal_lifelines} further confirms that broad scene correspondence and spatial relationships are more consistently interpreted by AI than other localized facility-, sign-, and landmark-based indicators, where specific indicators are consistently challenging no matter in term of years or lifelines. As a result, these heatmaps further answers the aforementioned question of when and where shall we trust volunteers over AI models.

%----------------------------------------------
%----------------------------------------------

%----------------------------------------------
%----------------------------------------------

For PhotoMappers, this phenomena has a direct geographic interpretation. Spatial relationships among buildings, roads, vegetation, and open space describe the configuration of a place and can remain informative even when individual objects change, while Landmark Cues, by contrast, depend more strongly on localized and place-specific features which are better captured by human volunteers than VLMs. However, what is ironic is that, in disaster scenes, precisely these features may be damaged, obscured, displaced, or visually transformed. In this context, VLMs may therefore recognize the broader scene and its spatial organization while remaining uncertain about the evidence that uniquely identifies the place.

%----------------------------------------------------------------
\subsubsection{Where and When AI Struggles}

In a nutshell, the objective of this longitudinal analysis is to identify where automated disaster photos interpretation using GeoAI and VLMs remains difficult. As one can see, AI is more consistent when geographic inference can be supported by broad scene characteristics and spatial layout, but becomes less reliable when localization depends on landmark cues and other localized or disrupted evidence. This difficulty varies across years, disaster types, and Community Lifelines because each context presents a different combination of geographic cues and disaster impacts.

At the same time, the persistent gains from R@1 to R@5 and R@10 demonstrate that failure at the top-ranked location does not necessarily imply failure to recover useful geographic information. GeoAI can often narrow a large search space to a small set of plausible candidates even where it struggles to resolve the final location. The reasoning results help explain this gap as broad spatial configurations are comparatively stable, whereas landmarks, signs, facilities, and other place-specific cues remain more difficult to interpret consistently.

These key findings offer a good basis to rethink about human--AI complementarity, similar to the volunteer-machine collaboration in \citet{Herfort2019} and \citet{huck_centaur_2021}. For PhotoMappers, automated workflow with AI can reduce the scale of geographic search and organize candidate evidence, while human mappers perform significantly better in cases where cues are ambiguous, damaged, or place-specific. 

To this end, volunteer's geographic knowledge and judgment remains particularly valuable for integrating multiple weak signals, incorporating event-specific context, and rejecting plausible but insufficiently supported matches. The operational opportunity is therefore not simply to automate disaster mapping process, but to use AI to concentrate human attention where geographic ambiguity is greatest and automated reasoning is least reliable.

%================================================================

\section{Discussion}
\label{sec:discussion}

Our study uses the decade-long PhotoMappers archive to examine how the geography of crowdsourced disaster information changes through time and how contemporary AI models performs within this changing context. By combining cross-view geolocalization with indicator-based evaluation of geographic reasoning, we show that model performance varies across years, disaster types, Community Lifelines, and geographic evidences. However, these findings should be interpreted with several limitations: first, PhotoMappers is an event-driven operational archive rather than a representative sample of disaster occurrence; second, some categories are sparsely populated; third, the complexity of geographic reasoning is necessarily simplified through the structured 30 indicators. Finally, the reference imagery has its own blind spots~\citep{2025_ceus_svi_coverage}.  Rather than considering these limitations only as constraints, we use them to extend our discussion to broader questions of crowdsourcing, human-AI interaction and responsible governance of GeoAI in disaster mapping.

%----------------------------------------------------------------
\subsection{A Changing Geography of Crowdsourced Disaster Mapping}

PhotoMappers demonstrates that disaster photos are produced through a distinctly crowdsourcing process. Its spatial distribution reflects not only where disasters occur, but also where photographs are available, where volunteers participate, and where emergency agencies activate the workflow. The resulting archive is therefore a geography of observed and mobilized disaster information rather than a facet map of disaster exposure. This extends longstanding concerns in VGI research about uneven participation, institutional priorities, and the unequal visibility of places and communities~\citep{haworth2018,burns2018}.

This observation matters for GeoAI because changing data composition changes the task itself. Different years contain different hazards, landscapes, infrastructure, and operational priorities, and therefore different configurations of geographic evidences. Temporal variation in model performance should consequently not be read simply as technological progress or decline. It reflects an interaction between model capability and the geography represented in the data. In this sense, spatial heterogeneity is not a outlier to be averaged away, but part of what it means for GeoAI to operate in the world.

In the US context, this unevenness is not merely an artifact but reflects deeper socio-spatial inequalities regarding digital access, infrastructure baselines, and which communities are prioritized by institutional mapping activations. Just as \citet{SHELTON2014167} demonstrated the unequal ''data shadows'' of VGI during Hurricane Sandy, automated GeoAI systems risk widening the digital divide if they implicitly optimize for well-mapped, highly visible urban centers while underperforming in rural or historically marginalized areas where crowdsourced data is sparse \citep{huang2025digitaldivide}. Therefore, relying solely on automated VLMs workflows could inadvertently funnel emergency resources toward visually unambigious regions while further marginalizing data-poor communities. The spatial heterogeneity of the PhotoMappers archive must be understood as part of what it means for GeoAI to operate within the unequal socio-geographic reality of the US.

%----------------------------------------------------------------
\subsection{Accuracy, Reasoning, and Beyond}

Our findings show why accuracy alone provides an incomplete evaluation of advanced GeoAI models. The persistent gap between R@1 and R@5/R@10 indicates that GeoAI can often recover a plausible geographic context without resolving the final location. The reasoning analysis adds another layer that VLMs are more consistent in recognizing broad spatial configurations than in interpreting distinctive, place-specific cues. In this context, accurate geolocalization and logical geographic reasoning are therefore related but not equivalent tasks.

This distinction is especially important in disaster mapping settings. Roads, buildings, vegetation, and other geographic relationships may remain interpretable at the scene level, while signs, facilities, landmarks, and other distinctive cues may be obscured or transformed by the disaster. A model can consequently produce a plausible location without providing reliable evidence for why that location should be preferred. Evaluating GeoAI thus requires attention to the geographic basis of predictions, not only their final coordinates. This complements recent work emphasizing spatial heterogeneity, uncertainty, and reproducibility as fundamental properties of GeoAI rather than secondary evaluation considerations~\citep{li2024geoai,li2026unbox}.

For geography, the broader implication is a shift from asking whether AI can locate places toward asking how AI constructs geographic knowledge. Such a shift makes reasoning, uncertainty, and contextual validity central evaluation criteria rather than post hoc additions to accuracy metrics.

%----------------------------------------------------------------
\subsection{Why VGI is still Important}

The emergence of increasingly capable VLMs does not make VGI less important. It changes the role of VGI from simply supplying more data to providing situated observations and volunteer geographical interpretation. During disasters, crowdsourced photos can capture ground conditions that are rapidly changing or temporarily inaccessible to conventional observation. Recent work in \citet{zhou2026rapid} similarly demonstrates how social-media data can support rapid disaster response while requiring attention to data access, privacy, and spatially uneven participation.

PhotoMappers highlights that the value of VGI lies in the process through which information is produced. Volunteers do not merely contribute photos, they search, interpret, geolocate, and validate them within an operational and dynamic workflow. This makes the resulting geographic information different from raw image observations. Hybrid approaches such as \citet{Herfort2019} and Centaur VGI~\citep{huck_centaur_2021} already pointed out that machine assistance can increase mapping speed and reduce volunteer burden without eliminating purposeful human participation.

This human component remains important because AI cannot compensate for persistent gaps in geographic observations. Places that are less photographed, less connected, or less likely to trigger volunteer participation may remain underrepresented even when machine processing becomes highly efficient. Indeed, recent work also warns that automated systems can deepen existing inequalities in data access, computational capacity, and geographic representation~\citep{huang2025digitaldivide}. VGI therefore remains not only a data source for GeoAI, but an important mechanism through which local observations and situated geographic knowledge enter disaster information systems.

These findings align directly with modern US defense and intelligence doctrines surrounding Human-Machine Teaming (HMT) \citep{palmer_accelerating_2024}. Within the US Intelligence Community, frameworks such as the Office of the Director of National Intelligence's Augmenting Intelligence Machines (AIM) Initiative explicitly emphasize using AI to augment, rather than replace, human analysis \citep{odni_aim_2019}. Operationally, this translates to the established GEOINT practice of 'tipping and cueing', where automated systems process massive, unstructured datasets to 'tip' off human analysts and 'cue' them to high-priority areas of interests \citep{biltgen_activity_2016}. Because our results show VLMs consistently recover the correct geographic context within a small candidate set based on broad spatial layouts, they can effectively reduce the geographic search space as an initial computational 'tip'. Human volunteers acting as citizen GEOINT analysts can then apply their spatial cognition and local knowledge to resolve the localized landmark cues that VLMs struggle with. 

%----------------------------------------------------------------
\subsection{How to Ensure Responsible GeoAI}

Responsible GeoAI cannot be reduced to higher model accuracy or a fixed set of technical safeguards. A recent framework proposed by \citet{li2026unbox} highlights four interconnected dimensions, Representativeness, Explainability, Sustainability, and Ethics, across Data, Application, and Society scopes. Rather than providing a one-for-all solution, these dimensions offer a useful lens for considering how GeoAI might be introduced more cautiously into operational disaster mapping.

\textbf{Representativeness} is particularly relevant to our findings. PhotoMappers is geographically and temporally uneven, and model performance varies across years, disaster types, and Community Lifelines. Aggregate accuracy may therefore be biased contexts in which evidence is sparse or model behavior is less reliable. Reporting disaggregated performance and explicitly identifying poorly represented cases can provide a more reliable evidence basis for deciding when GeoAI results require additional human auditing.

\textbf{Explainability} is equally important because a predicted coordinate alone provides limited information about why a location was selected. Our results suggest that VLMs are more consistent with broad spatial configurations than with localized Landmark Cues. In practice, an AI-assisted disaster mapping system shall therefore return several candidate locations together with the VGI-SVI-RSI evidence supporting them, rather than presenting a single location as deterministic. Cases with weak or conflicting evidence could then be directed to human review in a later stage.

\textbf{Sustainability and Ethics} introduce considerations that extend beyond model performance. GeoAI and VLMs carry computational costs, suggesting value in using lightweight retrieval for routine geolocalization and reserving more intensive reasoning for ambiguous observations. Moreover, disaster photos may also contain identifiable people, private property, or sensitive infrastructure. How such imagery is collected, retained, processed, and communicated therefore requires attention alongside improvements in model performance along.

For PhotoMappers, these considerations suggest that responsible GeoAI may be most useful when embedded within, rather than substituted for, the existing human validation workflow. AI could assist triage, propose candidate locations, and organize geographic evidence, while uncertain or consequential cases remain subject to human review before entering operational products. Such an arrangement does not eliminate error or guarantee responsible outcomes, but it can make uncertainty, provenance, and human accountability more visible within the mapping process.

Last but not least, responsible GeoAI is not only a question of whether AI is accurate, but also where its evidence comes from, where it remains uncertain, who validates its outputs, and what consequences follow when it is wrong. In short, accuracy asks whether AI is right; geographic reasoning asks why; responsible GeoAI asks who is represented, who validates it, and what happens when it is wrong.

%These considerations also align with emerging US federal mandates. A recent White House Executive Order on the Safe, Secure and Trustworthy Development and Use of Artificial Intelligence \citep{white_house_executive_2023} alongside the NIST AI Risk Management Framework \citep{tabassi_artificial_2023} explicitly demand human-in-the-loop safeguards, accountability and explainability when AI is deployed in high-consequence environments like critical infrastructure and disaster response. Returning a single, opaque VLM-generated coordinate in a life-safety context violates these principles. Our proposed approach where AI proposes candidate locations and organizes evidence, while human mappers audit ambiguous cases provides a practical, operational pathway for agencies to harness GeoAI while complying with federal mandates for algorithmic accountability.

%================================================================
\section{Conclusions}
\label{sec:conclusion}
%================================================================

This study used the decade-long PhotoMappers archive to examine the changing geography of crowdsourced disaster mapping and the capacity of GeoAI to geolocate and reason about real-world disaster imagery. Three findings stand out from the longitudinal analysis presented. First, crowdsourced disaster information is geographically and temporally uneven, reflecting changing hazards, places, Community Lifelines, and operational priorities; Second, integrating VGI, SVI, and RSI improves cross-view geolocalization, but performance varies substantially across years and disaster contexts; Third, geolocalization and geographic reasoning are related but distinct: GeoAI can often retrieve the correct location among a small set of candidates, yet remains less consistent when reasoning depends on localized and place-specific cues.

These key findings suggest that the future of disaster GeoAI lies in complementing rather than replacing volunteer geographic interpretation. AI can accelerate image triage, candidate retrieval, and the organization of geographic evidence, while human mappers remain important for resolving ambiguity, interpreting disrupted or place-specific cues, and validating consequential decisions. This approach directly supports modern operational doctrines of Human-Machine Teaming, positioning Vision-Language Models as advanced "tipping and cueing" systems that scale spatial search while leaving localized spatial judgment to citizen GEOINT analysts. More broadly, evaluating GeoAI requires moving beyond aggregate accuracy toward geographic context, reasoning, uncertainty, representation, and algorithmic accountability. For disaster response, the pressing need is therefore not simply to automate mapping, but to build responsible Human-AI-in-the-loop workflows that combine computational scale with crowdsourced geographic interpretation while complying with emerging federal mandates for safe and accountable AI.

\section*{Acknowledgments}

This work is supported by
(1) the Start-Up Grant (SUG) project “Geospatial Artificial Intelligence for Climate Resilient Urban Environment” from the National University of Singapore (E-109-00-0036-01); and
(2) the project "Assessing Urban Flood Resilience against Climate Extreme with GeoAI in Southeast Asia", which is funded by the Singapore Ministry of Education Academic Research Fund Tier 1 (A-8004907-00-00).

\section*{Disclosure statement}
The authors report there are no competing interests to declare.

\bibliographystyle{tfv}
\bibliography{interacttfvsample}

\end{document}